\documentclass{article}

\usepackage{arxiv}
\usepackage[T1]{fontenc}
\usepackage[utf8]{inputenc}

\usepackage{microtype}
\usepackage{graphicx}
\usepackage{subcaption}
\usepackage{wrapfig}
\newcommand{\ablationplot}[3]{
    \begin{wrapfigure}{r}{0.40\linewidth}
        \vspace{-\intextsep}
        \centering
        \captionsetup{font=small,skip=5pt}
        #1
        \caption{#2}\label{#3}
    \end{wrapfigure}}
\newcommand{\ablationresultplot}[4]{%
    \par\noindent
    \begin{minipage}[t]{0.58\linewidth}
        \vspace{0pt}
        #4
    \end{minipage}\hfill
    \begin{minipage}[t]{0.40\linewidth}
        \vspace{0pt}
        \centering
        \captionsetup{type=figure,font=small,skip=5pt,justification=raggedright}
        #1
        \caption{#2}\label{#3}
    \end{minipage}\par}
\usepackage{booktabs}
\usepackage[authoryear,round]{natbib}
\usepackage{hyperref}
\usepackage{xurl}
\usepackage{amsmath}
\usepackage{amssymb}
\usepackage{bm}
\usepackage{mathtools}
\usepackage{amsthm}
\usepackage[capitalize,noabbrev]{cleveref}
\usepackage{xspace}
\usepackage{multirow}
\usepackage{xcolor}
\usepackage{array}
\usepackage{placeins}

\newtheorem{theorem}{Theorem}
\newtheorem{corollary}{Corollary}
\newtheorem{assumption}{Assumption}
\newtheorem{proposition}{Proposition}

\newsavebox{\displaymathbox}
\newcommand{\fitmath}[2]{%
  \begingroup
  \sbox{\displaymathbox}{$\displaystyle #1$}%
  \ifdim\wd\displaymathbox>\linewidth
    \def\selectedmath{#2}%
  \else
    \def\selectedmath{#1}%
  \fi
  \expandafter\endgroup\selectedmath
}

\usepackage{tikz}
\usetikzlibrary{svg.path}

\definecolor{vitaTeal}{HTML}{065E58}
\definecolor{waveTeal}{HTML}{06766E}
\definecolor{vitaBlue}{HTML}{167195}
\definecolor{vitaIndigo}{HTML}{234C96}
\newcommand{\projectmark}{\tikz[baseline=(current bounding box.south),scale=.23,yscale=-1]{
  \path[fill=waveTeal] svg {M1 18C9 17 13 2 22 2C32 2 37 21 46 21C53 21 58 12 63 9C58 16 54 26 46 26C35 26 30 7 22 7C14 7 9 20 1 18Z};
  \path[fill=vitaBlue] svg {M1 28C9 27 13 12 22 12C32 12 37 31 46 31C53 31 58 22 63 19C58 26 54 36 46 36C35 36 30 17 22 17C14 17 9 30 1 28Z};
  \path[fill=vitaIndigo] svg {M1 38C9 37 13 22 22 22C32 22 37 41 46 41C53 41 58 32 63 29C58 36 54 46 46 46C35 46 30 27 22 27C14 27 9 40 1 38Z};
}}
\newcommand{\projecturl}{https://ucd-dare.github.io/VITA-WAM}
\newcommand{\equalmark}{\kern.06em\raisebox{.14ex}{\bfseries *}}
\newcommand{\corrmark}{\kern.06em\raisebox{.2ex}{\bfseries\textdagger}}
\newcommand{\papertitle}{VITA-WAM: Manifold-Aligned World-Action Modeling for Real-Time Robot Control}
\title{\papertitle}
\author{Boqi Zhao, Dechen Gao, Hanchu Zhou, Hang Wang, Junshan Zhang}
\date{}
\renewcommand{\shorttitle}{VITA-WAM}
\hypersetup{hidelinks,pdftitle={\papertitle},
  pdfauthor={Boqi Zhao, Dechen Gao, Hanchu Zhou, Hang Wang, Junshan Zhang},
  pdfkeywords={robot learning, flow matching policy, world action models, visuomotor control}}

\makeatletter
\let\arxivnormalsize\normalsize
\renewcommand{\normalsize}{\arxivnormalsize\@setfontsize\normalsize{11}{14}}
\makeatother
\AtBeginDocument{\normalsize\raggedbottom}
\renewcommand{\maketitle}{
  \thispagestyle{empty}
  \begingroup
  \centering\rmfamily\setlength{\parskip}{0pt}
  {\bfseries\fontsize{34}{38}\selectfont\textcolor{vitaTeal}{VITA-WAM}\par}
  \vspace{5pt}
  {\bfseries\fontsize{21}{24}\selectfont
   Manifold-Aligned World-Action Modeling\\for Real-Time Robot Control\par}
  \vspace{16pt}
  {\bfseries\fontsize{13}{16}\selectfont
    \href{https://anchor1021.github.io/}{Boqi Zhao}\equalmark\quad
    \href{https://gaodechen.github.io/}{Dechen Gao}\equalmark\quad
    \href{https://hanchuzhou.github.io/}{Hanchu Zhou}\quad
    \href{https://ustcmike.github.io/}{Hang Wang}\quad
    \href{https://faculty.engineering.ucdavis.edu/jzhang/}{Junshan Zhang}\corrmark\par
    \vspace{6pt}
    \mdseries\equalmark\enspace Equal contributors\qquad
    \corrmark\enspace Corresponding author\par}
  \vspace{10pt}
  {\fontsize{11.5}{14}\selectfont\color{vitaTeal}
    \href{\projecturl}{\raisebox{-1.3pt}{\projectmark}\enspace Project page: ucd-dare.github.io/VITA-WAM}\par}
  \endgroup
  \vspace{13pt}
}

\begin{document}
\maketitle

{\setlength{\leftmargini}{0pt}
\begin{abstract}
World action model (WAM) is a promising paradigm for visuomotor control by jointly modeling world dynamics and actions. Notably, existing WAMs often rely on large generative models and iterative sampling  which incur substantial latency for action generation,   limiting real-time deployment. In particular, there are two root causes of inefficiency in conventional flow matching or diffusion based action models in  WAMs: 1) transporting uninformative Gaussian noise into precise actions necessitating multiple ODE steps, and 2) repeatedly injecting task context through expensive cross-attention modules.  Our key insight is that jointly learning and aligning the source and action manifolds streamlines the transport between them, substantially accelerating action generation.
Building on this insight, we introduce VITA-WAM, a WAM architecture with manifold-aligned flow-matching (VITA flow matching) that directly transports latent observation representations containing visual, language, and proprioceptive information into latent actions. More specifically, a flow latent alignment (FLA) mechanism is employed to facilitate manifold alignment between latent observation and action representations, enabling accurate generation with few ODE steps. The resulting architecture also eliminates costly attention for observation conditioning during repeated denoising steps. Notably, with a single ODE step, VITA-WAM outperforms 10-step FastWAM by 21.47 pp on LIBERO-Plus and by 1.65 pp on LIBERO, achieving these gains through the FLA objective without distillation, while inferring at 55.67~ms latency, a $2.10\times$ speedup compared to FastWAM.
\end{abstract}

}

\newcommand{\ManifoldFigureOnFirstPage}{}
\begin{minipage}{\linewidth}
  \captionsetup{type=figure,font=small,skip=4pt}
  \centering
\includegraphics[width=\linewidth]{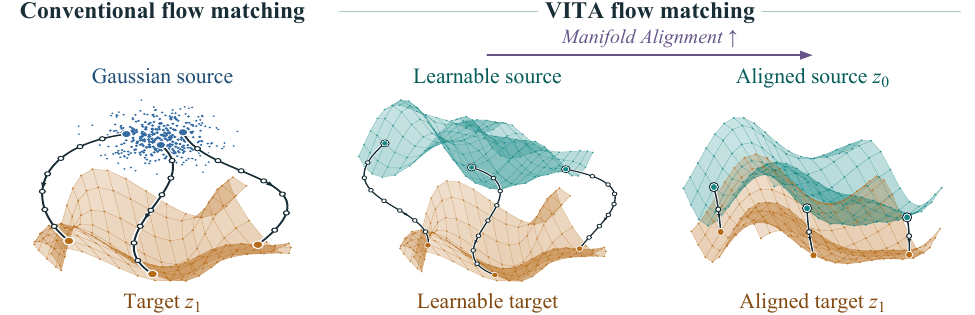}
\caption{Comparing conventional flow matching  (left) vs.~VITA flow matching (right). Conventional flow matching learns to transport Gaussian noise to the target manifold, necessitating multiple ODE steps. VITA flow matching (right) jointly learns and aligns the source and target representation manifolds with the Flow Latent Alignment (FLA) mechanism, thus enabling generation with simplified transport paths and fewer ODE steps. VITA-WAM employs VITA flow matching, significantly improving the efficiency of WAM action generation.}
\label{fig:manifold_alignment}

\end{minipage}

\section{Introduction}
\label{sec:introduction}

World Action Models (WAMs) have emerged as a promising paradigm for robot learning by jointly modeling physical dynamics and robot actions \citep{zhu2025uwm, li2025uva, bi2026motus, ye2026dreamzero}. Learning to predict future dynamics encourages representations that capture how the environment evolves, supporting physical understanding and policy generalization. Notably, existing WAMs often rely on large diffusion or flow matching based video and action models that require iterative sampling. The resulting latency during inference limits robots' ability to incorporate new observations and respond promptly to environmental changes for real-time control. A central challenge is therefore to reduce the computational cost of action generation while retaining the benefits of world modeling.

Recent work has made attempts to address this challenge. Fast-WAM eliminates explicit future video generation at inference while retaining video co-training to learn representations that enhance action prediction \citep{yuan2026fastwam}. Nevertheless, its action model uses conventional flow matching, with multiple ODE integration steps, repeating cross-attention conditioning on task observations at each step. Action generation still remains a computational bottleneck. Complementary approaches reduce the number of sampling steps through distillation or specialized few-step training objectives~\citep{geng2025meanflow,ye2026dreamzero,akbari2026flashwam}, which may  incur costly training procedures or even degrade performance.

A primary objective of this work is to overcome the inefficiency of action generation for WAMs, while maintaining the performance (e.g., success rate). Our earlier work, VITA flow matching~\citep{vita-arxiv,gao2026vita}, identified the sampling from uninformative Gaussian noise as a potential bottleneck to low-latency action generation. More specifically, conventional flow matching begins with a Gaussian prior as the source of the flow and progressively transforms it into actions, which is inherently inefficient for two reasons: 1) Complex Gaussian-to-action transport. Gaussian noise is uninformative and distant from the target action manifolds, necessitating multiple ODE steps for denoising into precise actions; 2) Heavy computational cost associated with cross-attention based conditioning. When starting from uninformative Gaussian sources, conventional flow-matching models must incorporate task contexts through heavy cross-attention based conditioning at each ODE step.

VITA addresses these inefficiencies by jointly learning the source and target latent manifolds and directly modeling the flow from observation latents to action latents. To the best of our knowledge, VITA~\citep{vita-arxiv} is the first work to apply a non-Gaussian source to flow matching policies, and to develop learnable source and target latents for flow matching. VITA proposed two flow latent objectives, flow latent decoding (FLD) and flow latent consistency (FLC), to enable source--target co-adaptation while preventing latent collapse arising from the training--inference discrepancy, in the sense that during training the target latent is provided by an action encoder, whereas at inference the action latent is generated by solving the ODE using the learned velocity field. The resulting joint optimization promotes alignment between the source and target manifolds, simplifying their transport and facilitating accurate action generation with fewer ODE steps. Moreover, since the source is grounded by observations, VITA eliminates the need for repeated conditioning modules.

Nevertheless, VITA studied FLD and FLC mainly through empirical studies, and is task-specific, constrained by a smaller-scale network of around 30M parameters without language conditioning. In this work, we develop VITA-WAM, scaling up VITA to a 6B WAM architecture, formalize flow latent objectives under a unified Flow Latent Alignment (FLA) mechanism, and establish a theoretic foundation for manifold alignment. A key challenge is to learn an informative flow source from visual, language, and proprioceptive representation while matching the dimensionality of latent actions, as required by flow matching.
To address this, we develop a Q-Former-based source encoder to effectively extract and fuse multimodal observations into a  compact, learnable source representation. 
The FLA mechanism, designed to minimize the reconstruction loss using ODE-generated latent actions, effectively boils down to the co-optimization of flow matching, and the source and target manifolds in VITA-WAM, thereby driving the source observations and target actions towards aligned manifolds. Our theoretic study shows that aligning the source and target latent representations reduces the coupling cost, facilitating simpler, less curved flow trajectories, and allowing the corresponding ODE to be solved accurately with fewer steps, thereby reducing inference latency, as illustrated in \Cref{fig:manifold_alignment}. From an optimization perspective, this joint learning shifts part of the action generation burden from the flow matching model to the source encoder, which learns an action-relevant initialization that simplifies the subsequent transport to latent actions.
{\em In essence, through FLA-enabled manifold alignment, VITA-WAM streamlines the flow between source latent and action latent, and speeds up the action generation process. }

\ifdefined\ManifoldFigureOnFirstPage\else
\begin{figure}[t]
    \input{figures/manifold_alignment}
\end{figure}
\fi

We conducted comprehensive empirical evaluations across four real-world tasks using a Franka Research 3 robot arm, as well as on the LIBERO and LIBERO-Plus simulation benchmarks. With a single ODE step, VITA-WAM outperforms 10-step FastWAM by 21.47 pp on LIBERO-Plus and 1.65 pp on LIBERO. On LIBERO-Plus, this improvement stems from FLA learning action-relevant representations that are invariant to visual perturbations, thereby improving robustness. With a single ODE step and repeated language and proprioception cross-attention removed, VITA-WAM achieves an inference latency of 55.67~ms, compared to 116.65~ms for 10-step FastWAM and 62.32~ms for 1-step FastWAM. We emphasize that these results are achieved without distillation.

Our contributions are:

1) \textbf{VITA-WAM: A highly efficient WAM architecture.} We introduce VITA-WAM, a novel world-action model for visuomotor control that substantially reduces action-generation latency. By jointly optimizing the source and target manifold representations with flow matching, VITA-WAM begins generation from an informative, learnable source and avoids repeated attention to multi-modal observations during denoising.
In particular, VITA-WAM learns manifold-aligned source and target latent representations for flow matching, simplifying transport and enabling low-latency action generation for real-time robot control.


2) \textbf{Flow Latent Alignment (FLA)}. We formalize FLA that enables VITA flow learning, and provide theoretic studies to demonstrate that FLA enforces the source and target manifold alignment of flow matching, thus simplifying the transport from the source to the target and reducing the ODE steps needed. The VITA flow using FLA is a general flow matching method that can be applied to the WAM or VLA architecture.

3) \textbf{State-of-the-art success rates and efficiency.} We conduct empirical studies on both real-world and simulation tasks. With a single ODE step, VITA-WAM achieves higher success rates than 10-step FastWAM while providing an $2.1\times$ inference speedup.

\section{Related Work}
\label{sec:related_work}

\paragraph{Efficient World Action Models.}
WAMs have shown great promise in visuomotor control by jointly modeling world dynamics and actions~\citep{feng2025vidar,pai2025mimicvideo,li2026lingbot,zhu2025uwm,bi2026motus,ye2026dreamzero,yuan2026fastwam}. Many state-of-the-art WAMs build on large video foundation models and iteratively generate video and actions, increasing inference latency and memory requirements and limiting real-time deployment. Fast-WAM~\citep{yuan2026fastwam} eliminates explicit future generation at inference. Several methods develop compact or predictive representations to reduce inference cost~\citep{li2026efficientwam,zhao2026fasterwamfuture,chen2026lawam,wang2026mowam,zhang2026imagewam}, and explore methods that simplify the action model~\citep{li2026lightwam,ma2026fasterwam} or reduce denoising steps through distillation~\citep{akbari2026flashwam,ye2026dreamzero}. A promising direction is to reconsider the source of flow matching itself~\citep{vita-arxiv,jia2026a2a,dai2026leap,zhou2026agile}, to simplify the underlying transport required for action generation, which is discussed below:

\textbf{Flow Source Design for Flow Matching Policies.} Conventional flow matching policies flow from Gaussian sources to actions~\citep{lipman2023flow,chi2025diffusion,zhang2025flowpolicy}. VITA flow matching~\citep{vita-arxiv} and Streaming Flow Policy~\citep{jiang2025streamingflow} pioneer the design of flow sources for flow matching policies. Departing from the conventional flow matching paradigm, VITA is the first non-Gaussian flow matching policy to jointly learn and align latent sources and latent targets using FLA, while removing repeated conditioning during ODE solving to improve efficiency. VITA can use a deterministic flow that speeds up convergence and improves control precision. However, VITA is task-specific and limited by its small network scale, and FLA has only been studied empirically.  Streaming Flow Policy uses historical actions to sample from an informed Gaussian source to facilitate flow matching. A2A (Action-to-Action)~\citep{jia2026a2a} uses historical actions as the source, while adopting a non-Gaussian source and latent action formulation of FLA. However, when temporal smoothness degrades, the history prior can introduce a strong bias that could hurt performance when adjacent action chunks are dissimilar; additionally, these methods still require costly conditioning. VITA-WAM applies VITA flow matching to action models in WAMs and inherits the noise-free and conditioning-free formulation from VITA. Furthermore, VITA-WAM scales to a 6B WAM, formalizes FLA, and establishes the theoretical foundation for manifold alignment.

\section{VITA-WAM: Manifold-Aligned World-Action Modeling}
\label{sec:method}

\subsection{Preliminaries}

\paragraph{Visuomotor Policy}
The action model in WAMs can be formulated as a visuomotor policy $\pi(A\mid O)$ that predicts action chunks given multimodal observations~\citep{black2025realtimeexecutionactionchunking}. The policy is  learned using behavioral cloning~\citep{torabi2018behavioral} on a demonstration dataset $\mathcal{D}=\{(O,A)\}$, where $O=(V,L,P)$ comprises current visual observations $V$, a language instruction $L$, and the proprioceptive state $P$. An action chunk is denoted by $A=a_{1:T_{\mathrm{pred}}}\in\mathbb R^{T_{\mathrm{pred}}\times D_{\mathrm{action}}}$, where $T_{\mathrm{pred}}$ is the prediction horizon and $D_{\mathrm{action}}$ is the action dimensionality.

\paragraph{Flow Matching~\citep{lipman2023flow}}
Flow matching learns a velocity field $v_\theta$ that transports source samples $\bm{z}_0\sim p_0$ to target samples $\bm{z}_1\sim p_1$. For paired sources and targets, the linear interpolation $\bm{z}_t=(1-t)\bm{z}_0+t\bm{z}_1$ defines a target velocity $\mathrm{d}\bm{z}_t/\mathrm{d}t=\bm{z}_1-\bm{z}_0$. The training objective of a flow matching model minimizes
\[
\mathcal{L}_{\mathrm{FM}}
=
\mathbb{E}_{\bm{z}_0,\bm{z}_1,t}
\left[
\left\|v_\theta(\bm{z}_t,t)-(\bm{z}_1-\bm{z}_0)\right\|_2^2
\right],
\]
where $t\sim\mathcal{U}[0,1]$. At inference, a numerical solver (e.g., the Euler method) integrates $\mathrm{d}\tilde{\bm{z}}_t/\mathrm{d}t=v_\theta(\tilde{\bm{z}}_t,t)$ from $\tilde{\bm{z}}_0=\bm{z}_0$ to $t=1$, yielding an approximate target sample $\hat{\bm{z}}_1$.

Conventional flow matching policies typically use a Gaussian source $p_0=\mathcal{N}(0,I)$ and an observation-conditioned velocity field $v_\theta(\bm{z}_t,t\mid O)$ during both training and inference.

\subsection{VITA Flow Matching~\citep{vita-arxiv,gao2026vita}}
\label{sec:vita_fm_method}

VITA flows directly from a learned source $\bm{z}_0=\mathcal E_v(V,P)$ encoding visual observations and proprioception to a latent action target $\bm{z}_1=\mathcal E_a(A)$. Since task observations are encoded in the source, the velocity field $v_\theta(\bm{z}_t,t)$ removes extra observation conditioning.
Flow matching requires the source $\bm{z}_0$ and target $\bm{z}_1$ to have the same dimensionality. VITA therefore introduces an action autoencoder to map raw actions into a latent space with the same dimensionality as the source representations, as discussed below.

\textbf{Action Autoencoder (AE).}
We learn the latent action space using an action autoencoder. The action encoder learns to encode raw action chunks into latent actions, and the action decoder learns to reconstruct raw action chunks from latent actions $\bm{z}_1$. The AE loss can be written as:
\[
\mathcal L_{\mathrm{AE}}
=
\mathbb E_{(O,A)\sim\mathcal D}\!\left[
    \mathcal L_{\mathrm{act}}\!\left(
        \mathcal D_a(\bm{z}_1),A
    \right)
\right].
\]

Both latent spaces are learned jointly with the flow using the flow matching objective defined above. At inference, numerical integration from $\bm{z}_0$ yields $\hat{\bm{z}}_1=\mathcal T(\bm{z}_0)$, which is decoded as $\hat A=\mathcal D_a(\hat{\bm{z}}_1)$. Furthermore, VITA introduces Flow Latent Alignment (FLA) to address latent collapse during joint training, discussed below:

\textbf{Flow Latent Alignment (FLA).}
FLA addresses the latent collapse caused by the training-inference mismatch in decoder inputs~\citep{gao2026vita}: during policy training, latent actions are produced by the action encoder, whereas at inference they are generated by solving the flow ODE. FLA directly supervises latent actions generated by ODE at training time to anchor flow ODE using raw action supervision:
\[
\mathcal L_{\mathrm{FLA}}
=
\mathbb E_{(O,A)\sim\mathcal D}\!\left[
    \mathcal L_{\mathrm{act}}\!\left(
        \mathcal D_a(\mathcal T(\bm{z}_0)),A
    \right)
\right],
\]
where $\mathcal L_{\mathrm{act}}(\hat A,A)=\|\hat A-A\|_1/D$ is the $L_1$ reconstruction loss normalized by the latent dimension $D$. Gradients propagate through the action decoder, numerical solver, and source encoder.

\textbf{VITA Loss Function.}
Combining AE, FLA, and flow matching losses, the overall objective for VITA flow matching is
\[
\mathcal L
=
\lambda_{\mathrm{FM}}\mathcal L_{\mathrm{FM}}
+\lambda_{\mathrm{FLA}}\mathcal L_{\mathrm{FLA}}
+\lambda_{\mathrm{AE}}\mathcal L_{\mathrm{AE}},
\]
where the $\lambda_{\mathrm{FM}}$, $\lambda_{\mathrm{FLA}}$, and $\lambda_{\mathrm{AE}}$ are coefficients to the three loss components.

\subsection{VITA-WAM Architecture}
\label{sec:vita_wam_architecture}

\paragraph{Architecture.}

\begin{figure*}[t]
    \centering
    \includegraphics[
        width=\textwidth,
        height=0.40\textheight,
        keepaspectratio
    ]{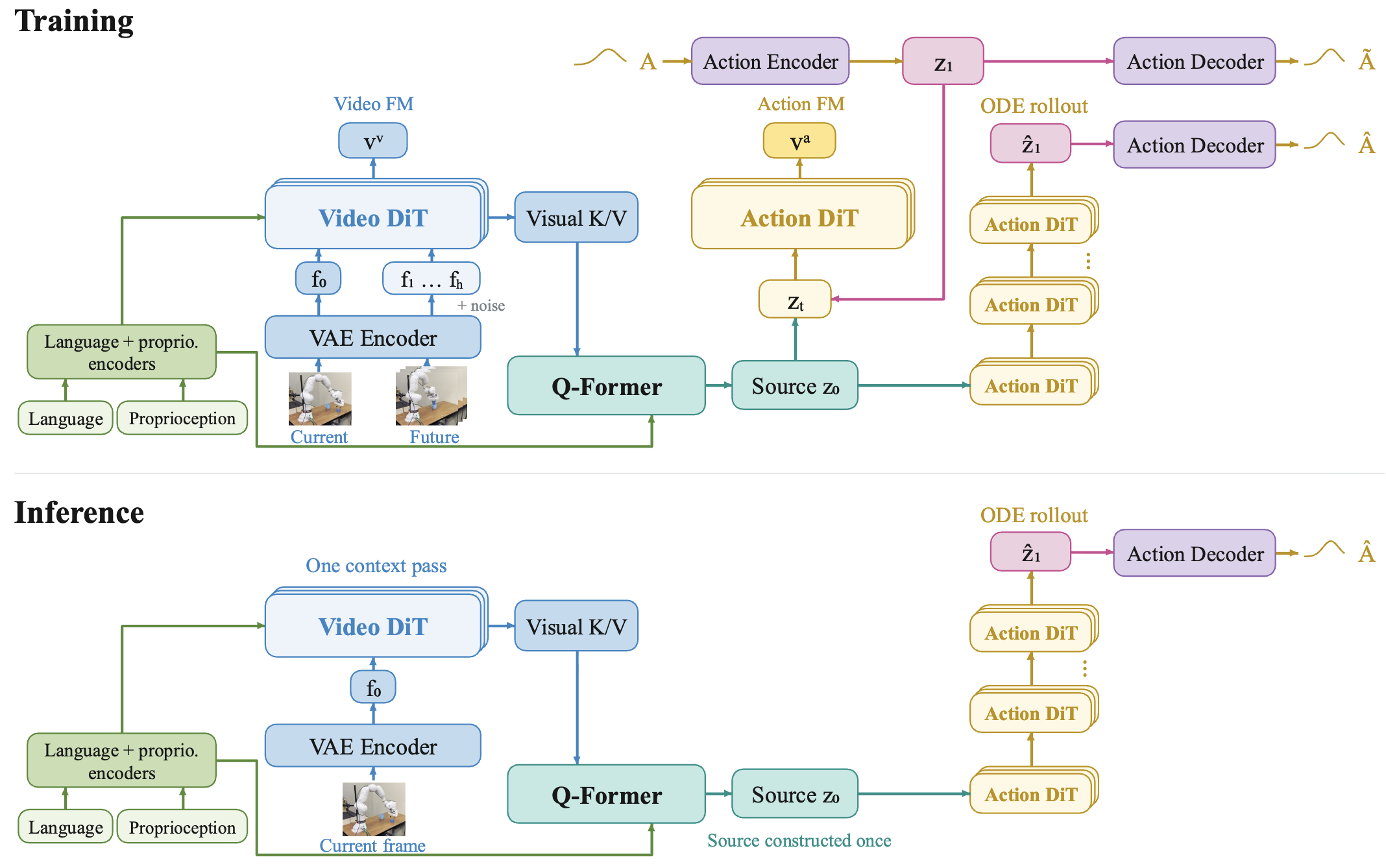}
    \captionsetup{labelfont=normalfont}
    \caption{Illustration of VITA-WAM architecture. 1) During training (top),  a Q-Former fuses visual, language and proprioception into a latent source $z_0$, where the visual signal is the output of the video DiT supervised by video flow matching. The action encoder maps the action chunk into a target latent $z_1$. The source and target representations are jointly learned with VITA flow matching through FM, AE and FLA objectives, with FLA supervising the action chunk decoded from the ODE-generated latent.
    2) During inference (bottom), a single video DiT forward pass extracts current-observation features, which are fused with language and proprioception by the Q-Former to construct the source latent $z_0$. The action DiT then performs conditioning-free ODE integration starting from $z_0$ to generate the action latent $\hat{z}_1$, which is decoded into an action chunk.}
    \label{fig:vita_wam_training_inferece}
\end{figure*}

VITA-WAM is distinguished from prior WAM architectures in its way of incorporating multi-modal observations and predicting the flow. Existing WAMs such as FastWAM use an action DiT for action generation which samples from Gaussian noise for flow matching, and their action DiT attends to video, language, and proprioceptive tokens through costly attention at each flow step.
In contrast, VITA-WAM replaces the Gaussian source sampling with a learnable source latent that fuses visual features, language embeddings, and proprioceptive states into $\bm{z}_0$ with a Q-Former~\citep{li2023blip2}. Since task information has been encoded in $\bm{z}_0$, VITA-WAM removes the action model's attention to multi-modal observations. The action model therefore operates only on the evolving action latent, avoiding repeated attention to observation features at each flow step. The action decoder decodes the generated latent actions $\hat{\bm{z}}_1$ to an action chunk.

More specifically, as illustrated in Fig.~\ref{fig:vita_wam_training_inferece},  VITA-WAM builds upon a video diffusion transformer (DiT) architecture, using a pretrained Wan2.2-TI2V-5B video DiT~\citep{wan2025} following prior WAMs~\citep{bi2026motus, li2026lingbot, yuan2026fastwam}. Language instructions are encoded by a T5 text encoder~\citep{2020t5}, while proprioceptive states are linearly projected into a token appended to the language embeddings. Visual tokens are produced by the video DiT.

\textbf{Learning Sources and Latent Action Targets for VITA-WAM.} Flow matching requires the source and target to have identical dimensionalities. However, multimodal observations generally have much higher dimensionality than action representations. We therefore learn a compact source latent $\bm{z}_0$ to encode observations, and upsample action chunks to target latent actions $\bm{z}_1$ of the same dimensionality. Specifically, the source encoder $\mathcal{E}_s$ uses a trainable Q-Former that queries and fuses multimodal inputs $\bm{z}_0=\mathcal{E}_s(O)$, while the action encoder $\mathcal{E}_a$ maps action chunks to target latents $\bm{z}_1=\mathcal{E}_a(A)$, where $\bm{z}_0,\bm{z}_1\in\mathbb R^D$.

\textbf{Training.} VITA-WAM trains video DiT using video flow matching and action DiT using VITA flow matching. Video flow matching is standard in existing WAMs. The VITA flow matching objectives contain three components, the AE, FLA, and flow matching losses discussed above in~\Cref{sec:vita_fm_method}.

\textbf{Inference}. The Q-Former encodes observations into $\bm{z}_0$, which is transported to latent action $\bm{z}_1$ without extra observation conditioning through the flow ODE. An action decoder $\mathcal{D}_a$ maps latent actions to an action chunk. The policy is therefore $\hat A=\pi(O)=\mathcal{D}_a\!\left(\mathcal T(\bm{z}_0)\right)$, where $\mathcal T$ denotes the numerical flow map from $t=0$ to $t=1$.

\section{Flow Latent Alignment (FLA): Towards Aligning Source Manifold and Target Manifold for Efficient Flow Matching}

In this section, we first formalize Flow Latent Alignment (FLA) in VITA flow matching. We then present a  theoretical framework on manifold alignment between the latent source and target manifold, and show rigorously how FLA reduces flow curvature and thus enables efficient flow matching.

\begin{figure*}[!htbp]
    \centering
    \includegraphics[width=\textwidth]{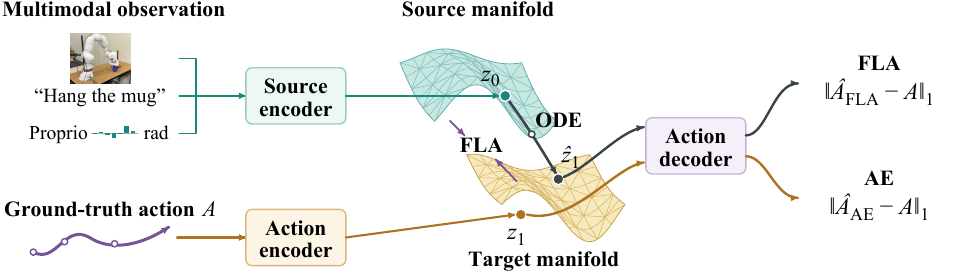}
    \caption{Illustration of Flow Latent Alignment (FLA) in VITA-WAM}
    \label{fla-manifold}
\end{figure*}
\FloatBarrier

As illustrated in Fig.~\ref{fla-manifold}, Flow latent alignment (FLA) closes the training-inference gap by supervising the ODE's own output. During training,  we integrate the flow from the source latent $\bm z_0=\mathcal E_v(O)$ with the same $K$-step ODE solver used at inference, $\hat{\bm z}_1=\mathcal T_\theta (\bm z_0)$, decode the result, and compare it with the ground-truth action:

$$
\mathcal L_{\mathrm{FLA}}
=
\mathbb E_{(O,A)\sim\mathcal D}\!\left[
\mathcal L_{\mathrm{act}}\!\big(
\mathcal D_a(\mathcal T_\theta (\bm z_0)),\,A
\big)
\right],
\qquad
\mathcal L_{\mathrm{act}}(\hat A,A)=\frac{1}{D}\,\|\hat A-A\|_1 ,
$$

where $\mathcal T_\theta(\bm z_0)=\bm z_0+\Delta t\sum_{k=0}^{K-1}v_\theta(\bm z_{t_k},t_k)$ is the   $K$-step Euler map and $D$ is the action dimension times chunk length. Gradients propagate through the action decoder, every solver step (and hence the flow network $v_\theta$), and the source encoder $\mathcal E_v$. The flow is therefore trained on the states it visits at test time, with raw action supervision that does not pass through the target latent. The target encoder $\mathcal E_a$ lies off this path and is shaped by the flow-matching and autoencoding terms.

\subsection{Manifold Alignment between Source Manifold and Target Manifold}
\label{sec:manifold-alignment}

In general, manifold alignment is the process of learning ``projection functions'' from distinct data manifolds into a shared latent space   that simultaneously bring corresponding points across manifolds close together while preserving the intrinsic local geometry and neighborhood topology of each individual manifold. In VITA-WAM,  let  $Z_0:=\mathcal E_s(O)$ and $Z_1 : =\mathcal E_a(A)$ be the two mappings from the source and target to the latent space, respectively, and both $Z_0$ and $Z_1$ are  random variables in the latent space $\mathbb R^D$, corresponding to the sample latents $\bm{z}_0$ and $\bm{z}_1$. Here $D$ is the latent dimension and $O=(V,L,P)$ includes visual, language, and proprioceptive inputs. Throughout this section, task-context conditioning is held fixed and left implicit in velocity fields, conditional expectations, variances, and laws (distributions); outer expectations average over task contexts and data with model parameters fixed, and $p_i=\mathrm{Law}(Z_i)$. By definition, for given $\mathcal E_s, \mathcal E_a$,  manifold alignment  between multi-modal  observations and actions,  is measured by
\[
\fitmath{
 \mathcal C(\mathcal E_s,\mathcal E_a):=\mathbb E_{(O,A)}\big\|\mathcal E_a(A)-\mathcal E_s(O)\big\|^2
 =\mathbb E\|Z_1-Z_0\|^2\Big|_{\text{training coupling}\ \gamma}
}{
 \begin{aligned}
 \mathcal C(\mathcal E_s,\mathcal E_a)&:=\mathbb E_{(O,A)}\big\|\mathcal E_a(A)-\mathcal E_s(O)\big\|^2\\
 &=\mathbb E\|Z_1-Z_0\|^2\Big|_{\text{coupling}\ \gamma}.
 \end{aligned}
}
\]
Formally, manifold alignment aims to find  the optimal coupling $\gamma \in \Pi(p_0,p_1) $ to minimize
$\mathcal C(\mathcal E_s, \mathcal E_a)$.
It is known that the squared Wasserstein distance,  
$$W_2(p_0,p_1)^2 = \min_{\gamma\in\Pi(p_0,p_1)}\mathbb E_\gamma\|Z_0-Z_1\|^2$$
is  a lower bound on the transport cost of any coupling, and is attained when the  coupling itself is  the optimal transport (OT) one.
That is to say, when the  pairing $(Z_0,Z_1)$ has the OT coupling, $\mathbb E\|Z_1-Z_0\|^2=W_2(p_0,p_1)^2$.
For ease of reference, we define the  coupling excess corresponding to the coupling   resulted from training as
$$
T_e = \mathcal C(\mathcal E_s, \mathcal E_a)  - W_2(p_0,p_1)^2
$$

For convenience, let $Z_t=(1-t)Z_0+tZ_1$ and define $r_t:=v_\theta(Z_t,t)-(Z_1-Z_0)$. Let $\widetilde Z_1=\mathcal T(Z_0)$ be the generated endpoint by solving the ODE and define $u:=\widetilde Z_1-Z_0$, so $\widetilde Z_1=Z_0+u$.
Let $\bar u$ denote the average velocity $\int_0^1v_\theta(Z_t,t)\,dt$, and define $\rho:=\int_0^1 r_t\,dt$ and $\delta:=Z_1-\widetilde Z_1$. 
Then, the training-inference gap can be measured by $\|\bar u-u\|_{L^2}$.

For a random vector $X$, write $\|X\|_{L^2}:=(\mathbb E\|X\|_2^2)^{1/2}$ and $\|X\|_{L^1}:=\mathbb E\|X\|_1$.
As introduced in Section~3, we use squared $L_2$ error for flow matching (FM),  and $L_1$ reconstruction losses for FLA (flow latent alignment) and AE (action encoder). With $D$ the latent dimension, define the per-sample reconstruction losses
\[
\fitmath{
 \ell_{\mathrm{FLA}}:=\frac{\|\mathcal D_a(\widetilde Z_1)-A\|_1}{D},\qquad
 \ell_{\mathrm{AE}}:=\frac{\|\mathcal D_a(Z_1)-A\|_1}{D}.
}{
 \begin{gathered}
 \ell_{\mathrm{FLA}}:=\frac{\|\mathcal D_a(\widetilde Z_1)-A\|_1}{D},\\
 \ell_{\mathrm{AE}}:=\frac{\|\mathcal D_a(Z_1)-A\|_1}{D}.
 \end{gathered}
}
\]
It follows that the training losses are
\begin{align*}
  \mathcal L_{\mathrm{FM}}^{\mathrm a}&=\int_0^1\mathbb E\|r_t\|_2^2\,dt
  = \mathbb{E} 
\left\|v_\theta(\bm{z}_t,t)-(\bm{z}_1-\bm{z}_0)\right\|_2^2
  , \\
 \mathcal L_{\mathrm{FLA}}&=\mathbb E\ell_{\mathrm{FLA}}
 =\frac{\|\mathcal D_a(\widehat Z_1)-A\|_{L^1}}{D},\ &
  \mathcal L_{\mathrm{AE}} =\mathbb E\ell_{\mathrm{AE}}
 =\frac{\|\mathcal D_a(Z_1)-A\|_{L^1}}{D}.
\end{align*}
The normalization constant  $1/D$ can be absorbed into the reconstruction-loss weights. As in the method, timestep weighting is omitted from the displayed FM loss.

\begin{assumption}
\label{assump:decoder-regularity}
$\mathcal{D}_a$ is continuously differentiable ($C^1$) in the action latent space, with Jacobian singular values bounded as $m\!\le\!\sigma_{\min}\!\le\!\sigma_{\max}\!\le\!L$ in the neighborhood.
\end{assumption}

We have the  first main result on the relationship between VITA-WAM's loss functions and manifold alignment between the source and target.

\begin{theorem}
\label{thm:ma-upperbound}
Under Assumption~\ref{assump:decoder-regularity}, manifold alignment between multi-modal observations and actions satisfies
\[
\fitmath{
 \sqrt{\mathcal C(\mathcal E_s,\mathcal E_a)}\le\min\!\left\{
 \|\bar u\|_{L^2}+\sqrt{\mathcal L_{\mathrm{FM}}^{\mathrm a}},\;
 \|u\|_{L^2}+\frac{D}{m}\|\ell_{\mathrm{FLA}}+\ell_{\mathrm{AE}}\|_{L^2}\right\}.
}{
 \begin{aligned}
 \sqrt{\mathcal C(\mathcal E_s,\mathcal E_a)}
 &\le\min\!\left\{\|\bar u\|_{L^2}+\sqrt{\mathcal L_{\mathrm{FM}}^{\mathrm a}},\right.\\
 &\hspace{2em}\left.\|u\|_{L^2}+\frac{D}{m}\|\ell_{\mathrm{FLA}}+\ell_{\mathrm{AE}}\|_{L^2}\right\}.
 \end{aligned}
}
\]
\end{theorem}
The proof is given in Appendix~\ref{app:l1-proofs}.

We have the following engineering insight based on Theorem~\ref{thm:ma-upperbound}.

\textbf{1) Action-centric representation of multi-modal observations.} FLA jointly learns the source and target manifolds along with the flow matching model. 
In particular, it drives the multi-modal source encoder (Q-Former) and target action encoder to cooperate towards manifold alignment: $\mathcal E_s$ must keep the action-relevant content of $O$ in $Z_0$, and $\mathcal E_a$ must place $Z_1$ where $Z_0$ can predict it. In this way, Q-Former (multi-modal source encoder) learns action-centric representation of the multi-modal inputs.

\textbf{2) Reduced training-inference gap.} $\mathcal L_{\mathrm{FLA}}$ runs the ODE from $Z_0=\mathcal E_s(O)$ and requires the rollout endpoint to decode to the true action. Among all functions of $Z_0$ at fixed task context, the conditional mean is the best mean-squared predictor of $Z_1$, so
\[
\fitmath{
 \mathbb E\|Z_1-\mathbb E[Z_1\mid Z_0]\|_2^2
 \le\frac{D^2}{m^2}\mathbb E(\ell_{\mathrm{FLA}}+\ell_{\mathrm{AE}})^2.
}{
 \begin{aligned}
 \mathbb E\|Z_1-\mathbb E[Z_1\mid Z_0]\|_2^2
 &\le\mathbb E\|Z_1-\widehat Z_1\|_2^2\\
 &\le\frac{D^2}{m^2}\mathbb E(\ell_{\mathrm{FLA}}+\ell_{\mathrm{AE}})^2.
 \end{aligned}
}
\]
The left side is $\mathbb E\,\mathrm{Var}(Z_1\mid Z_0)$, the amount of the ambiguity in flow matching. Thus small $L_1$ reconstruction errors in the   second-moment sense above limit ambiguity at the start of the flow, driving both encoders to cooperate towards manifold alignment.
The combined losses therefore shape both encoders: $\mathcal E_s$ is shaped to make $Z_0$ carry the action-relevant content of the observation, and $\mathcal E_a$ is shaped to place action latents where the flow can reach them. Moreover, $\mathcal L_{\mathrm{AE}}$ encourages the target latent $Z_1=\mathcal E_a(A)$ to remain decodable with high fidelity.
The training--inference gap satisfies
\[
 \|\bar u-u\|_{L^2} \le\sqrt{\mathcal L_{\mathrm{FM}}^{\mathrm a}}
 +\frac{D}{m}\|\ell_{\mathrm{FLA}}+\ell_{\mathrm{AE}}\|_{L^2}.
\]

\subsection{Manifold Alignment, Source-Target Transport Cost, and Flow Curvature} 

For convenience, let  $v_{\theta}(Z, t)$ denote the trained model (parameterized by $\theta$) in the real world, and 
$ v^*(x,t)=\mathbb E[ Z_1 - Z_0\mid Z_t=x]$.
In what follows, we first study the ideal case  with $v_\theta = v^*$, assuming finite second moments and a well-posed ODE that preserves the reference marginals at each fixed task context.


For convenience, let   $\{\widetilde Z_t \}$ denote the ODE process induced by $\dot{\widetilde Z}_t=v^*(\widetilde Z_t,t)$ with initial condition $\widetilde Z_0:=Z_0$. Let $\widetilde Z_1$ denote its terminal value, which we caution is not the same random variable as the original latent target $Z_1$.
Rewrite the ODE process explicitly, 
$$\dot{\widetilde Z}_t= v^*({\widetilde Z}_t,t),\qquad {\widetilde Z}_0:=Z_0$$
Let $ \mathcal S(\widetilde{\mathbf Z})$ denote the  path curvature of the above ODE path:
\[
\mathcal S(\widetilde{\mathbf Z}):=\int_0^1\mathbb E\big\|\dot{\widetilde Z}_t-(\widetilde Z_1-\widetilde Z_0)\big\|^2dt
\]

We have the following result on the source-target coupling cost for the case with $v_\theta = v^*$.

\begin{proposition} [The case with $v_\theta = v^*$]
\label{thm:couplingcost1}
The coupling cost between the source latent $Z_0$ and target latent $Z_1$ can be decomposed as
\begin{equation*}
\mathbb E\|Z_1-Z_0\|^2 = \underbrace{\mathbb E\|\widetilde Z_1- Z_0\|^2}_{\text{realized transport}} + \underbrace{\mathcal S(\widetilde{\mathbf Z})}_{\text{curvature}} + \underbrace{\mathcal L_{\mathrm{FM}}^{\mathrm a}(v^*)}_{\text{irreducible FM floor}}
\end{equation*}
\end{proposition}

We emphasize that the coupling  cost depends on the joint law, and the two pairs
$(Z_1, Z_0)$ and $(\widetilde Z_0,\widetilde Z_1)$ have different joint laws in general. More specifically,
$$\mathbb E\|Z_1-Z_0\|^2=\mathbb E\|Z_0\|^2+\mathbb E\|Z_1\|^2-2\, \mathbb E\langle Z_0,Z_1\rangle.$$  
where the first two terms depend only on the marginals. If $\mathrm{Law}(\widetilde Z_0)=\mathrm{Law}(Z_0)$ and $\mathrm{Law}(\widetilde Z_1)=\mathrm{Law}(Z_1)$, they are the same for both pairs. The  coupling cost may  differ only through the cross term, which is a property of the joint law. Subtracting the two expansions,
$$ \mathbb E\|Z_1-Z_0\|^2  - 
\mathbb E\|\widetilde Z_1-\widetilde Z_0\|^2 
=2\Big(\mathbb E\langle\widetilde Z_0,\widetilde Z_1\rangle-\mathbb E\langle Z_0,Z_1\rangle\Big).$$
Note that the ODE-induced coupling's realized transport cost is generally a cheaper, less-crossing pairing.

 Next, we  examine the source-target coupling cost    $ \mathbb E\|Z_1-Z_0\|^2 = \|Z_1-Z_0\|^2_{L^2}  $,   for the case with $v_\theta \ne v^*$.      We use $\widetilde Z_t^{\theta}$ for the rollout driven by $v_\theta$, i.e., $\dot{\widetilde Z^{\theta}_t} =v_\theta(\widetilde Z^{\theta}_t,t)$, $\widetilde Z_0^\theta=Z_0$.
 Recall that $\dot{\widetilde Z}_t=v^*(\widetilde Z_t,t)$ with initial condition $\widetilde Z_0=Z_0$.
 By marginal preservation, we have that $\mathrm{Law}(\widetilde Z_t)=\mathrm{Law}(Z_t)$ at each fixed task context, for all $t$. It is clear that
$\widetilde Z_t$ and $\widetilde Z_t^{\theta}$ are different rollouts  path-wise.
By the triangle inequality,
\[ 
\|    Z_1 - Z_0\|_{L^2} \leq  \|     Z_1 - \widetilde Z_1^{\theta}\|_{L^2} + \underbrace{\|  \widetilde Z_1^{\theta}
-Z_0\|_{L^2}}_{\text{realized transport under  $v_\theta$}}
\]


\begin{assumption} 
\label{assump:ode}


 Assume $v_\theta(\cdot,t)$  is $\Lambda$-Lipschitz in $z$ for all $t$.  Define $\lambda=\frac{e^{2\Lambda}-1-2\Lambda}{4\Lambda^2}$  and $\bar\kappa=1+\Lambda\sqrt{\lambda}$. 
\end{assumption}


\begin{theorem}  
\label{thm:bound-align}
(a) Under Assumption 1, we have the following result on the flow  path curvature for the   case with $v_\theta = v^*$:
\begin{equation*}
  {\mathcal S(\widetilde{\mathbf Z})}  
  +  {\mathcal L_{\mathrm{FM}}^{\mathrm a}(v^*)} 
  \le \mathcal C(\mathcal E_s, \mathcal E_a)  - W_2(p_0,p_1)^2 
\end{equation*}

(b) Under Assumptions $1 \&  \ 2$, we have the following result on the flow  path curvature  for the   case with $v_\theta \ne v^*$:
$$  {\mathcal S(\widetilde{\mathbf Z}) 
+  \mathbb E  \|\widetilde Z^{\theta}_1 -Z_1\|^2 
\;\le\;\bar\kappa^2\big(  \mathcal C(\mathcal E_s, \mathcal E_a)  - W_2(p_0,p_1)^2 +\Delta_{FM}\big) 
}$$
with  $\Delta_{\mathrm{FM}}(\theta)=\mathcal L_{\mathrm{FM}}^{\mathrm a}(\theta) -\mathcal L_{\mathrm{FM}}^{\mathrm a}(v^*)$.

\end{theorem}

The detailed proofs of 
 Theorems~\ref{thm:ma-upperbound} and  \ref{thm:bound-align} can be found in the Appendix.




Combining Theorems~\ref{thm:ma-upperbound} and  \ref{thm:bound-align}, we conclude that under Assumptions 1 and 2, smaller FM loss and $L_1$ reconstruction errors   tighten the upper bound on the coupling cost $\mathbb E\|Z_1-Z_0\|^2$ for fixed transport terms and decoder conditioning.
Since  manifold alignment (measured by the source-target coupling cost) serves as an upper  bound on the sum of path curvature and regression ambiguity of the ideal flow, better alignment between the  source and target would shape the transport map toward an easy-to-represent  function, and hence improve the performance of multi-step unrolled training.
 The FLA mechanism in VITA-WAM serves as a regularizer to drive source-target latent manifold alignment,  by ``forcing'' the decoder to reconstruct real actions from both the encoder's output and the flow's rollout endpoint:  It implicitly pressures the geometry of the two latent distributions toward each other, not just fitting better functions on top of whatever geometry already exists. 
 When the source and target are well aligned, the coupling cost between $Z_1$ and $Z_0$ is small. Euler's global error is governed by how much the velocity changes along the trajectory, and this is intimately related to the variation that the path curvature term $\mathcal S(\widetilde{\mathbf Z})$ measures. A smaller coupling cost tightens the upper bound on path curvature, and a good flow matching model results in small $\mathcal L_{\mathrm{FM}}^{\mathrm a}(v^*)$, making it possible that using fewer Euler steps  in solving the ODE can work well and hence reduce the inference latency.

Summarizing, training VITA-WAM with the combined loss pushes the source and target latents toward a pairing that is easy to transport. $\mathcal L_{FLA}$ requires the rollout endpoint to decode to the true action, and the shared decoder  then pins that endpoint to the target latent $Z_1$, which  therefore must be nearly a function of the source latent $Z_0$, so the visual and world-model latent already carries the action-relevant content. $\mathcal L^a_{FM}$ shapes the encoders through its ambiguity floor, discouraging latent arrangements where many $(Z_0,Z_1)$ lines cross the same point. By  Theorems~\ref{thm:ma-upperbound} and  \ref{thm:bound-align}, at the ideal field the curvature and ambiguity are bounded by the pairing's suboptimality gap $C-W_2^2$, so a near-optimal, nearly deterministic pairing leaves little room for crossings or bends.

A flow that is nearly  unambiguous is cheap to integrate. In particular, Euler's error is governed by how much the velocity changes along the trajectory, which is the curvature, and a constant-velocity flow is integrated exactly in one step. So the number of ODE steps $K$ can drop, and since the steps are sequential evaluations of the flow network, action-generation latency falls roughly in proportion, with one decoder pass added. Training FLA at the same small $K$ used at inference gives a direct certificate that the small-step endpoint can be accurate. The practical result is a lower-latency control loop for real-time manipulation.

 
\section{Experiments}
\label{sec:experiments}


\subsection{Experimental Setup}
\label{sec:exp_setup}

\textbf{Simulation Benchmarks.} We evaluate VITA-WAM on LIBERO \citep{liu2023libero} and LIBERO-Plus \citep{fei2025liberoplus}. For LIBERO, we follow the standard benchmarking practice and train our models on the four LIBERO suites: Spatial, Object, Goal, and Long. Each suite contains 500 demonstrations spanning 10 tasks. We report success rates for each suite, evaluated over a total of 2000 episodes across 40 tasks. LIBERO-Plus is more challenging and tests robustness under distribution shifts across 10,030 cases that perturb object layouts, camera viewpoints, initial robot states, language instructions, lighting, background textures, and sensor noise. We directly evaluate the checkpoints trained on standard LIBERO without additional training or data augmentation.

\textbf{Real-World Benchmarks.} We conduct real-world experiments on a Franka Research 3 (FR3) robot arm as shown in~\Cref{fig:task_viz}. We consider four tasks: Pick Place, Stack Cups, Hang Mug, and Store Drawer. We collect 180 demonstrations per task, yielding 720 demonstrations in total (Appendix~\ref{app:real_world_data}). We report task success rates and inference latency per action chunk.



\textbf{Training \& Evaluation.} For LIBERO-40, we train both VITA-WAM and Fast-WAM for 20k steps with a global batch size of 128, evaluate checkpoints every 500 steps, and select the checkpoint with the highest average SRs across all suites in LIBERO-40. We evaluate each policy's selected checkpoint on LIBERO-Plus, evaluating all 10,030 cases.


For the real-world experiments, we train each policy on the 720 demonstrations from all four tasks for 30k steps with a global batch size of 128. We deploy the final checkpoints on the Franka Research 3 robot arm, using VITA-WAM ($K=1$) and Fast-WAM ($K=10$). We measure inference latency from a recorded observation and instruction to a 32-action chunk on a single NVIDIA RTX 4090 (Appendix~\ref{app:inference_latency}).

\subsection{Main Results}
\label{sec:main_results}

\paragraph{Simulation Performance.}
~\Cref{tab:libero40_results} and ~\Cref{tab:libero_plus_results} compare VITA-WAM's performance to state-of-the-art WAM and VLA methods on LIBERO-40 and LIBERO-Plus, respectively. VITA-WAM generates actions using $K=1$ or $K=2$ ODE steps, while the baselines are evaluated using their reported settings. For Fast-WAM, we additionally evaluate it with $K=1$ and $K=2$ to compare against VITA-WAM at the same ODE steps.

On LIBERO-40, VITA-WAM achieves overall success rates of 98.85\% at $K=1$ and 98.65\% at $K=2$ on LIBERO-40, exceeding Fast-WAM at matched step counts by 3.55 and 2.25 pp, and the default Fast-WAM at 10 ODE steps by 1.65 pp. On LIBERO-Plus, VITA-WAM achieves overall success rates of 70.80\% at $K=1$ and 73.05\% at $K=2$, outperforming Fast-WAM ($K=10$) by 21.47 and 23.72 pp, respectively. Using $K=2$ improves overall success by 2.25 pp and raises SRs in six of the seven perturbation categories, with the largest gains on Sensor Noise (+5.00 pp) and Light Conditions (+3.94 pp). The two-step model is within 0.52 pp of Faster-WAM (SparseMoT) using $K=10$.

\begin{table*}[!htbp]
    \centering
    \caption{LIBERO SRs (\%). $K$ is the number of ODE steps for flow matching. Bold values indicate the highest SRs.}
    \label{tab:libero40_results}
    \small
    \setlength{\tabcolsep}{4pt}
    \begin{tabular*}{\textwidth}{@{\extracolsep{\fill}}lcccccc@{}}
        \toprule
        Method & $K$ & Spatial & Object & Goal & Long & Overall \\
        \midrule
        $\pi_{0.5}$ & 10 & 98.80 & 98.20 & 98.00 & 92.40 & 96.85 \\
        LingBot-VA & $10$ & 98.50 & 99.60 & 97.20 & \textbf{98.50} & 98.50 \\
        Light-WAM & -- & 98.20 & 99.60 & 97.80 & 93.00 & 97.20 \\
        Faster-WAM (DoT, v2) & 10 & 99.10 & 99.50 & \textbf{98.40} & 96.20 & 98.30 \\
        Faster-WAM (SparseMoT) & 10 & 99.60 & \textbf{99.80} & 98.20 & 98.20 & \textbf{98.90} \\
        \midrule
        Fast-WAM & 1 & 96.20 & 99.00 & 95.40 & 90.60 & 95.30 \\
        Fast-WAM & 2 & 97.60 & 98.80 & 95.60 & 93.60 & 96.40 \\
        Fast-WAM & 10 & 98.20 & 99.60 & 97.60 & 93.40 & 97.20 \\
        \midrule
        \textbf{VITA-WAM} & 1 & \textbf{100.00} & 99.60 & 98.20 & 97.60 & 98.85 \\
        \textbf{VITA-WAM} & 2 & 98.40 & 99.60 & 98.20 & 98.40 & 98.65 \\
        \bottomrule
    \end{tabular*}
\end{table*}

\begin{table*}[!htbp]
    \centering
    \caption{LIBERO-Plus SRs (\%).}
    \label{tab:libero_plus_results}
    \footnotesize
    \setlength{\tabcolsep}{2pt}
    \begin{tabular*}{\textwidth}{@{\extracolsep{\fill}}lcccccccc@{}}
        \toprule
        Method & Camera & Robot & Lang. & Light & BG & Noise & Layout & Overall \\
        \midrule
        UniVLA & 1.80 & 46.20 & 69.60 & 69.00 & 81.00 & 21.20 & 31.90 & 42.90 \\
        OpenVLA-OFT & 56.40 & 31.90 & 79.50 & 88.70 & \textbf{93.30} & 75.80 & 74.20 & 69.60 \\
        $\pi_{0}$ & 13.80 & 6.00 & 58.80 & 85.00 & 81.40 & \textbf{79.00} & 68.90 & 53.60 \\
        $\pi_{0}$-Fast & \textbf{65.10} & 21.60 & 61.00 & 73.20 & 73.20 & 74.40 & 68.80 & 61.60 \\
        WorldVLA & 0.10 & 27.90 & 41.60 & 43.70 & 17.10 & 10.90 & 38.00 & 25.00 \\
        Fast-WAM (K=10) & 10.01 & 46.90 & 67.66 & 82.57 & 56.78 & 28.04 & 66.75 & 49.33 \\
        Faster-WAM (SparseMoT) (K=10) & 53.75 & \textbf{71.57} & \textbf{94.67} & \textbf{96.27} & 61.29 & 63.57 & 79.09 & \textbf{73.57} \\
        \textbf{VITA-WAM} ($K=1$) & 57.72 & 67.48 & 85.69 & 88.53 & 54.18 & 64.21 & 78.23 & 70.80 \\
        \textbf{VITA-WAM} ($K=2$) & 59.66 & 67.35 & 87.25 & 92.47 & 56.32 & 69.21 & \textbf{79.87} & 73.05 \\
        \bottomrule
    \end{tabular*}
\end{table*}

\paragraph{Real-World Performance and Efficiency.}
Across four tasks on the Franka Research 3, VITA-WAM ($K=1$) succeeds in 77 of 80 episodes (96.25\%), compared with 74 of 80 (92.50\%) for Fast-WAM ($K=10$), a gain of 3.75 pp (\Cref{tab:real_world_results}) while generating actions more efficiently. 

\Cref{tab:inference_efficiency} compares inference latency per action chunk in the real-world task setting on a single NVIDIA RTX 4090. VITA-WAM ($K=1$) generates an action chunk in 62.51~ms, compared with 211.41~ms for Fast-WAM ($K=10$), a $3.38\times$ speedup. At matched ODE step counts, VITA-WAM also has lower latency than Fast-WAM (62.51 versus 66.49~ms at $K=1$ and 71.07 versus 81.50~ms at $K=2$). With compilation, VITA-WAM ($K=1$) takes 55.67~ms per chunk, compared with 116.65~ms for Fast-WAM ($K=10$), a $2.10\times$ speedup. Faster-WAM (DoT), with a single-layer action DiT and 1.03B fewer checkpoint parameters than VITA-WAM, takes 62.56~ms per chunk before compilation, 0.05~ms longer than VITA-WAM ($K=1$). With compilation, DoT has the lowest measured latency at 55.28~ms, 0.39~ms lower than VITA-WAM ($K=1$).
Additionally, \Cref{tab:inference_efficiency} reports the number of floating-point operations (FLOPs) per action chunk. VITA-WAM ($K=1$) has the lowest estimated cost, at 2.6876~TFLOPs per chunk, 28.1\% lower than Fast-WAM ($K=10$).

\begin{table*}[!htbp]
    \centering
    \caption{Real-world task success rates (\%). Each policy is evaluated on 20 episodes per task.}
    \label{tab:real_world_results}
    \small
    \setlength{\tabcolsep}{4pt}
    \begin{tabular*}{\textwidth}{@{\extracolsep{\fill}}lccccc@{}}
        \toprule
        Method & \shortstack{Pick\\Place} & \shortstack{Stack\\Cups} & \shortstack{Hang\\Mug} & \shortstack{Store\\Drawer} & \shortstack{Overall\\SR (\%)} \\
        \midrule
        Fast-WAM ($K=10$) & \textbf{100} & 95 & 85 & 90 & 92.50 \\
        \textbf{VITA-WAM} ($K=1$) & 95 & \textbf{100} & \textbf{90} & \textbf{100} & \textbf{96.25} \\
        \bottomrule
    \end{tabular*}
\end{table*}

\begin{table*}[!htbp]
    \centering
    \caption{Comparison of inference efficiency. We report latency (ms) and computational cost (TFLOPs) per action chunk. All measurements are conducted on a single NVIDIA RTX 4090. $K$ denotes ODE steps used for flow matching. Appendix~\ref{app:inference_latency} details the timing protocol and parameter counts.}
    \label{tab:inference_efficiency}
    \footnotesize
    \setlength{\tabcolsep}{2pt}
    \begin{tabular*}{\textwidth}{@{\extracolsep{\fill}}lcccccc@{}}
        \toprule
        Method & $K$ & \shortstack{Action\\layers} & \shortstack{Latency\\(ms)$\downarrow$} & \shortstack{Latency\\(compiled, ms)$\downarrow$} & \shortstack{ TFLOPs\\per chunk} & \shortstack{Model \\Params (B)} \\
        \midrule
        $\pi_{0.5}$-DROID & 10 & 18 & 123.38 & 75.02 & 4.7530 & 3.62 \\

        Faster-WAM (SparseMoT) & 10 & 30 & 179.21 & 117.33 & 4.9236 & 5.56 \\
        Faster-WAM (DoT) & 10 & 1 & 62.56 & 55.28 & 2.6964 & 5.03 \\
        \midrule
        Fast-WAM & 1 & 30 & 66.49 & 62.32 & 2.7878 & 6.02 \\
        Fast-WAM & 2 & 30 & 81.50 & 62.99 & 2.8932 & 6.02 \\
        Fast-WAM & 10 & 30 & 211.41 & 116.65 & 3.7370 & 6.02 \\
        \midrule
        \textbf{VITA-WAM} & 1 & 30 & 62.51 & 55.67 & 2.6876 & 6.06 \\
        \textbf{VITA-WAM} & 2 & 30 & 71.07 & 58.06 & 2.6903 & 6.06 \\
        \bottomrule
    \end{tabular*}
\end{table*}

\FloatBarrier

\subsection{Ablation Studies}
\label{sec:ablations}

\paragraph{Ablation of Stochasticity in VITA Flow Matching Source}
VITA-WAM applies a VITA flow for action generation that starts from observation encoding, while prior work commonly considers the multi-modality enabled by stochasticity to be a key reason for the success of diffusion and flow matching policies. However, VITA shows that determinism can be favored in many robotics tasks to enhance precision, efficiency, and convergence while retaining high success rates. Much Ado about Noising~\citep{pan2026much} shows that multi-modality is not the primary reason for the success of flow matching policies, consistent with these findings.

\ablationplot{\includegraphics[width=\linewidth]{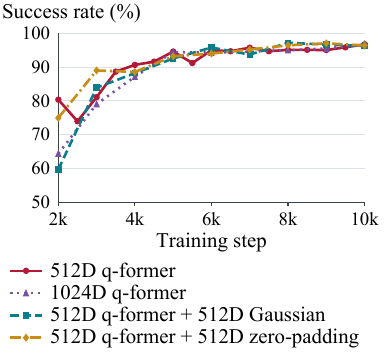}}{Ablation of stochasticity in the latent source on LIBERO.}{fig:source_construction_ablation}
We empirically study the effect of stochasticity in the VITA flow source and ablate the latent source construction. We first compare the default VITA-WAM, which uses a 512D latent source given by the Q-Former, with a variant that concatenates this source with a randomly sampled 512D Gaussian vector to introduce stochasticity. To control for the increased source dimensionality, we add a deterministic variant with a directly learned 1024D source. Since this also increases the source-encoder capacity, we further include a variant that pads the 512D source with 512D zeros, increasing dimensionality without adding information. The Gaussian and zero-padding variants share the same architecture and both encode observations in 512D, isolating the effect of the Gaussian noise. Figure~\ref{fig:source_construction_ablation} shows that the default deterministic source achieves 80.35\% success at 2k updates, compared with 59.65\% for the Gaussian variant. At 12k updates, all four variants reach similar success rates of 96.40--97.25\%. These results indicate that added source stochasticity provides no consistent advantage and that a deterministic action model is sufficient for high success rates on LIBERO.

\paragraph{Ablation of Q-Former}

Q-Former~\citep{li2023blip2} learns compact representations of high-dimensional multimodal observations, which can help capture information relevant to action generation~\citep{garrido2026learning}. In VITA-WAM, the Q-Former output serves as the source of the VITA flow. To test whether using this compact representation as conditioning can yield similar gains for the baseline, Fast-WAM, we use the same Q-Former to replace Fast-WAM's language and proprioception (LP) conditioning context. The original Fast-WAM, labeled LP (w/o Q-Former), conditions action generation directly on LP. LP (16 tokens) gives Q-Former the same LP inputs and uses its 16 output tokens as conditioning, testing the effect of learned fusion with the same conditioning modalities fixed. LPV (16 tokens) adds visual features to the inputs of Q-Former, while keeping the output at 16 tokens. LPV (1 token) keeps the same LPV inputs and aggregates the 16 query outputs into one token, which matches the design of VITA-WAM's source. All Q-Former output tokens are 512D. All four Fast-WAM variants retain the Gaussian action source and video-action MoT and use $K=10$ for action generation.

\ablationresultplot{\includegraphics[width=\linewidth]{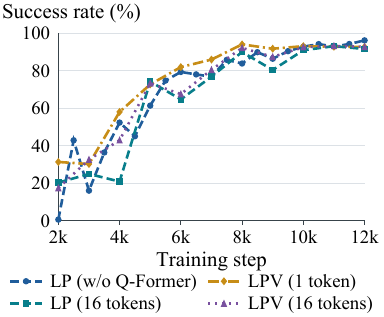}}
{Ablation of Q-Former on Fast-WAM to test the effect of using a compact observation representation.}
{fig:qformer_conditioning_ablation}
{At 8k training steps, LP (16 tokens), LPV (16 tokens), and LPV (1 token) reach 90.05\%, 92.25\%, and 94.10\% success, respectively, compared with 83.90\% for LP (w/o Q-Former) (\Cref{fig:qformer_conditioning_ablation}). Therefore, compact Q-Former conditioning improves early learning in Fast-WAM. When applying the same Q-Former representations for action generation, Fast-WAM LPV (1 token) reaches 92.75\% at 12k training steps with the token used as conditioning ($K=10$), whereas VITA-WAM reaches 97.50\% with the token used as the source of flow matching (see $K=1$ in \Cref{fig:libero_learning_curves}), demonstrating that WAMs benefit from both compactness of representations and using latent observation as the flow matching source.}

\FloatBarrier

\section{Conclusion}
\label{sec:conclusion}

We develop VITA-WAM, a novel world-action model for visuomotor control that substantially reduces action-generation latency. By jointly optimizing the source and target manifold representations with flow matching using a Flow Latent Alignment (FLA) objective, VITA-WAM learns aligned source and target manifold representations, reducing the number of ODE steps needed for generation. Furthermore, because the flow matching source is grounded by the multi-modal observations, VITA-WAM removes the costly conditioning module in the action model, further improving the efficiency. Additionally, we establish the theoretic foundation for manifold alignment, and demonstrate that FLA facilitates simpler and less curved flow trajectories. We conduct empirical studies on both simulation and real-world manipulation tasks. Compared to 10-step Fast-WAM, VITA-WAM improves success rates by 1.65 pp on LIBERO and 21.47 pp on LIBERO-Plus. When deployed on a single 4090 GPU, VITA-WAM infers at 55.67 ms latency without any additional distillation, achieving a $2.10\times$ speedup compared to Fast-WAM at the same model size of 6B parameters.

\bibliographystyle{plainnat}
\bibliography{references}

\clearpage
\appendix

\section{From VITA-WAM Training Loss to Manifold Alignment Between Source Latent and Target Latent}
\label{app:l1-proofs}

Recall that  $G:=Z_1-Z_0$,
$r_t:=v_\theta(Z_t,t)-G$, and $\rho:=\int_0^1r_t\,dt$.
Let $\widetilde Z_1=\mathcal T(Z_0)$ be the generated endpoint,
$u:=\widetilde Z_1-Z_0$, $\delta:=Z_1-\widetilde Z_1$, and
$\bar u:=\int_0^1v_\theta(Z_t,t)\,dt$.
For a random vector $X$, write
$\|X\|_{L^2}:=(\mathbb E\|X\|_2^2)^{1/2}$.
The identities $G=u+\delta=\bar u-\rho$ hold for the generation
procedure used in training and inference.

Write the per-example $L_1$ reconstruction errors as
\[
\fitmath{
 \ell_{\mathrm{FLA}}:=\frac{\|\mathcal D_a(\widetilde Z_1)-A\|_1}{D},\qquad
 \ell_{\mathrm{AE}}:=\frac{\|\mathcal D_a(Z_1)-A\|_1}{D}.
}{
 \begin{gathered}
 \ell_{\mathrm{FLA}}:=\frac{\|\mathcal D_a(\widetilde Z_1)-A\|_1}{D},\\
 \ell_{\mathrm{AE}}:=\frac{\|\mathcal D_a(Z_1)-A\|_1}{D}.
 \end{gathered}
}
\]
The training objectives are $\mathcal L_{\mathrm{FLA}}=\mathbb E [ \ell_{\mathrm{FLA}}] $
and $\mathcal L_{\mathrm{AE}}=\mathbb E [ \ell_{\mathrm{AE}}] $;
the quadratic bounds below use the second moment of these per-example
$L_1$ errors, which is distinct from their expected loss.
The FM loss remains
$\mathcal L_{\mathrm{FM}}^{\mathrm a}=\int_0^1\mathbb E\|r_t\|_2^2\,dt$.

By Assumption~\ref{assump:decoder-regularity}, the decoder's pairwise
lower bound and the triangle inequality through $A$ give, pointwise,
\begin{align}
 m\|\delta\|_2
 \le\|\mathcal D_a(Z_1)-\mathcal D_a(\widetilde Z_1)\|_2
 \le\|\mathcal D_a(Z_1)-\mathcal D_a(\widetilde Z_1)\|_1 \\ \nonumber
 \le\|\mathcal D_a(Z_1)-A\|_1+\|\mathcal D_a(\widetilde Z_1)-A\|_1
 =D(\ell_{\mathrm{FLA}}+\ell_{\mathrm{AE}}).
\end{align}
Taking $L^2$ norms on both sides  yields
\[
 \|\delta\|_{L^2}\le\frac Dm
 \|\ell_{\mathrm{FLA}}+\ell_{\mathrm{AE}}\|_{L^2}.
\]
Applying Cauchy--Schwarz's Inequality also gives
\[
 \|\rho\|_{L^2}^2
 =\mathbb E\left\|\int_0^1r_t\,dt\right\|_2^2
 \le\int_0^1\mathbb E\|r_t\|_2^2\,dt
 =\mathcal L_{\mathrm{FM}}^{\mathrm a}.
\]
Applying the triangle inequality to $G=u+\delta=\bar u-\rho$ proves
Theorem~\ref{thm:ma-upperbound}:
\[
\fitmath{
 \sqrt{\mathcal C}\le\min\!\left\{
 \|\bar u\|_{L^2}+\sqrt{\mathcal L_{\mathrm{FM}}^{\mathrm a}},\;
 \|u\|_{L^2}+\frac Dm\|\ell_{\mathrm{FLA}}+\ell_{\mathrm{AE}}\|_{L^2}\right\}.
}{
 \begin{aligned}
 \sqrt{\mathcal C}\le\min\!\Big\{&
 \|\bar u\|_{L^2}+\sqrt{\mathcal L_{\mathrm{FM}}^{\mathrm a}},\\
 &\|u\|_{L^2}+\frac Dm\|\ell_{\mathrm{FLA}}+\ell_{\mathrm{AE}}\|_{L^2}\Big\}.
 \end{aligned}
}
\]
Manifold alignment bounds curvature-plus-ambiguity through the coupling
training actually uses. The shared decoder connects the FLA and AE
reconstruction errors to agreement between the rollout endpoint and the
target latent, under the stated decoder-conditioning assumption.

{\bf Conditional ambiguity at the start of the flow.}
At fixed task context, $\widetilde Z_1=\mathcal T(Z_0)$ is a function of $Z_0$.
Since the conditional mean is the best mean-squared predictor,
\[
\fitmath{
 \mathbb E\,\mathrm{Var}(Z_1\mid Z_0)
 \le\mathbb E\|Z_1-\widetilde Z_1\|_2^2
 \le\frac{D^2}{m^2}\mathbb E(\ell_{\mathrm{FLA}}+\ell_{\mathrm{AE}})^2.
}{
 \begin{aligned}
 \mathbb E\,\mathrm{Var}(Z_1\mid Z_0)
 &\le\mathbb E\|Z_1-\widetilde Z_1\|_2^2\\
 &\le\frac{D^2}{m^2}\mathbb E(\ell_{\mathrm{FLA}}+\ell_{\mathrm{AE}})^2.
 \end{aligned}
}
\]
This controls ambiguity at $t=0$, rather than the time-integrated FM floor.

{\bf The training--inference gap.}
Subtracting the two identities gives $\bar u-u=\rho+\delta$.
Minkowski's inequality therefore yields
\[
 \|\bar u-u\|_{L^2}
 \le\sqrt{\mathcal L_{\mathrm{FM}}^{\mathrm a}}
 +\frac Dm\|\ell_{\mathrm{FLA}}+\ell_{\mathrm{AE}}\|_{L^2}.
\]
The left side is the distance between the flow's average velocity on the
training path and its displacement on its own rollout. FM alone need not
control this gap without additional stability assumptions:
Assumption~\ref{assump:ode} provides an exact-ODE endpoint bound, while
numerical rollouts also require control of solver error. FLA directly
supervises the generation procedure used at inference.

\section{Why Does Manifold Alignment Reduce Latency in Action Generation?}
\label{app:l2-proofs}

\subsection{A Lower Bound on Flow Matching Loss}

\paragraph{Notation.}
In this paper,  $O=(V,L,P)$; uppercase $Z$ denotes a random latent vector and lowercase
$\bm z$ a realization. Task-context conditioning is held fixed and left implicit in
velocity fields, conditional expectations, variances, and laws. Outer
expectations average over task contexts and data, $t$ is uniform on $[0,1]$,
and $\mathrm{Var}$ denotes scalar trace variance.
The weights of $\mathcal E_s,\mathcal E_a$ are $\phi_s,\phi_a$, with
$\phi=(\phi_s,\phi_a)$ and $\mathcal C(\phi)=\mathbb E\|Z_1-Z_0\|^2$.
$D$ is the latent dimension.
$\mathcal L_{\mathrm{FM}}^{\mathrm a}$ is the squared-$L_2$ flow-matching loss,
with uniform time and without scheduler weighting. 
The $\mathcal L_{\mathrm{FLA}}$ and
$\mathcal L_{\mathrm{AE}}$ objectives use $L_1$ reconstruction errors,
with the fixed normalization $1/D$.
The generated endpoint is $\widetilde Z_1=\mathcal T(Z_0)$.

Let $Z_0=\mathcal E_s(O)$ and $Z_1=\mathcal E_a(A)$.
Given a source latent distribution \(p_0\) and target latent distribution \(p_1\), flow matching learns a time-dependent vector field \(v_\theta(\bm z,t)\) whose flow transports
$
p_0 \longrightarrow p_1.$
To construct its training objective, one typically first selects a coupling
\[
\gamma\in\Pi(p_0,p_1),
\]
then defines a conditional probability path between paired endpoints, such as
\[
Z_t=(1-t)Z_0 + t Z_1.
\]
The target conditional velocity is then $
G =Z_1-Z_0, $
and the model minimizes
\[
\mathcal L_{\mathrm{FM}}^{\mathrm a}
=
\mathbb E_{\substack{(Z_0, Z_1)\sim\gamma\\t\sim\mathcal U[0,1]}}
\left[
\left\|v_\theta(Z_t,t)-G \right\|^2
\right].
\]

For given $Z_t$,  write $v(Z_t,t)=c$. Applying the bias--variance decomposition  pointwise at each $(Z_t,t)$,
we have 
$$\mathbb E\big[\|G-c\|^2 \mid Z_t=\bm z\big] = \underbrace{\mathrm{Var}(G\mid Z_t=\bm z)}_{\text{irreducible}} + \underbrace{\|\mathbb E[G\mid Z_t=\bm z]-  \widetilde G_t \|^2}_{\text{reducible, controlled by }c}$$

The second term is zero exactly when $ \widetilde G_t  =\mathbb E[G\mid Z_t=\bm z]=v^*(\bm z,t)$, and strictly positive otherwise.
Integrate over all $(\bm z,t)$. Since this holds pointwise, it holds after taking the outer expectation over $Z_t$:
\begin{equation}
 \mathcal L_{\mathrm{FM}}^{\mathrm a}(v) = \mathbb E_{t,Z_t}\Big[\mathrm{Var}(G\mid Z_t,t) + \|v^*(Z_t,t)-v(Z_t,t)\|^2\Big] \;\ge\; \mathbb E_{t,Z_t}\big[\mathrm{Var}(G\mid Z_t,t)\big] =: \mathcal L_{\mathrm{FM}}^{\mathrm a}(v^*)
\label{lv*}
\end{equation}
with equality iff $v=v^*$ almost everywhere. So $\mathcal L_{\mathrm{FM}}^{\mathrm a}(v^*)$ is  an irreducible flow matching loss of the underlying coupling geometry,  i.e., the intrinsic ambiguity exists at each point of the interpolated path, and no choice of $v$, however expressive, can push the loss below it. This is  the precise sense of a lower bound: it's the information lost by compressing $(Z_0,Z_1)$ down to $Z_t$ before predicting the slope.


{\bf Flow Matching Gap in the real-world case with $v_\theta\ne v^*$.} 
Let  $v_{\theta}(Z, t)$ denote the trained model in the real world. Write   $v^*(Z_t,t)=\mathbb E[G\mid Z_t]$, and $e_\theta(Z_t,t):=v_\theta(Z_t, t)-v^*(Z_t, t)$:
$$v_\theta(Z_t,t)-G = \underbrace{e_\theta(Z_t,t)}_{\text{function of }Z_t} + \underbrace{(v^*(Z_t,t)-G)}$$
We then have that 
$$\mathcal L_{\mathrm{FM}}^{\mathrm a}(\theta) = \underbrace{\int_0^1\mathbb E\|v_\theta(Z_t,t)-v^*(Z_t,t)\|^2\,dt}_{=:\Delta_{\mathrm{FM}}(\theta)\ \ge 0} + \mathcal L_{\mathrm{FM}}^{\mathrm a}(v^*)$$
because the cross term vanishes by the same orthogonality argument as before.  
Observe that  $\Delta_{\mathrm{FM}}(\theta)=\mathcal L_{\mathrm{FM}}^{\mathrm a}(\theta) -\mathcal L_{\mathrm{FM}}^{\mathrm a}(v^*)$ is the amount by which training flow matching loss exceeds its floor, and is exactly the squared $L^2$ distance between $v_\theta$ and $v^*$,  measured on the training distribution $Z_t$, the ground-truth interpolant.




\subsection{Decomposition of Source-Target Coupling Cost: The case with $v_\theta = v^*$.}

In what follows, we   first show that 
$\mathbb E\|Z_1-Z_0\|^2 = \int_0^1\mathbb E\|v^*(Z_t,t)\|^2\,dt + \mathcal L_{\mathrm{FM}}^{\mathrm a}(v^*)$ when $v_\theta = v^*$.

Intuitively, $\widetilde G_t$ is the orthogonal projection of $G$ onto the subspace of random variables that are functions of $Z_t$ at fixed task context.
Define the residual random variable
$R_t := G-\widetilde G_t = (Z_1-Z_0)-v^*(Z_t,t)$.
Clearly,  $\mathbb E[R_t\mid Z_t]=0$.  Since $\widetilde G_t=g(Z_t)$ for some function $g$ (namely $g=v^*(\cdot,t)$), use the tower property:
$$\mathbb E\big[R_t^\top\widetilde G_t\big] = \mathbb E\Big[\mathbb E[R_t^\top\widetilde G_t\mid Z_t]\Big] = \mathbb E\Big[\mathbb E[R_t\mid Z_t]^\top\widetilde G_t\Big] = \mathbb E\big[0^\top\widetilde G_t\big]=0$$
Expand the square, using $G=\widetilde G_t+R_t$:
$$\mathbb E\|G\|^2 = \mathbb E\|\widetilde G_t+R_t\|^2 = \mathbb E\|\widetilde G_t\|^2+2\,\mathbb E[\widetilde G_t^\top R_t]+\mathbb E\|R_t\|^2$$

Since the cross term vanishes, we have that  for every t,
$$\mathbb E\|Z_1-Z_0\|^2 \;=\; \mathbb E\|v^*(Z_t,t)\|^2 \;+\; \mathbb E\big\|Z_1-Z_0-v^*(Z_t,t)\big\|^2$$
This holds for every $t\in[0,1]$ separately. Integrate both sides over $t$:
$$\int_0^1\mathbb E\|Z_1-Z_0\|^2\,dt \;=\; \int_0^1\mathbb E\|v^*(Z_t,t)\|^2\,dt \;+\; \int_0^1\mathbb E\big\|Z_1-Z_0-v^*(Z_t,t)\big\|^2\,dt$$

The integrand on the left, $\mathbb E\|Z_1-Z_0\|^2$, does not depend on $t$;  $G=Z_1-Z_0$ is the same random vector regardless of $t$.  With the definition  of $\mathcal L_{\mathrm{FM}}^{\mathrm a}(v^*)$ in Eq.\ref{lv*}, we have proved
\begin{equation}  
\ \mathbb E\|Z_1-Z_0\|^2 \;=\; \int_0^1\mathbb E\|v^*(Z_t,t)\|^2\,dt \;+\; \mathcal L_{\mathrm{FM}}^{\mathrm a}(v^*)\
\label{energy}
\end{equation}

 Note that $v^*(Z_t,t)$ is  the conditional expectation  of $Z_1-Z_0$ for a given $Z_t$, used in the ODE corresponding to the velocity field.

{\bf ODE-induced process.}  $\widetilde Z_t$, defined by $\dot{\widetilde Z}_t=v^*(\widetilde Z_t,t)$ with initial condition $\widetilde Z_0:=Z_0$ (same starting random variable, but a genuinely different process from $t>0$ on). Let $\widetilde Z_1$ denote its terminal value, which we caution is not the same random variable as the original latent target $Z_1$.
Rewrite the ODE process explicitly, 
$$\dot{\widetilde Z}_t= v^*({\widetilde Z}_t,t),\qquad {\widetilde Z}_0:=Z_0$$



Since $\mathrm{Law}({\widetilde Z}_t)=\mathrm{Law}(Z_t)$ for every $t$, and the expectation of a function of a random variable depends only on its law, it follows that for any fixed $t$:
$$\mathbb E\big[\|v^*({\widetilde Z}_t,t)\|^2\big] = \mathbb E\big[\|v^*(Z_t,t)\|^2\big]$$
By construction $\dot {\widetilde Z}_t=v^*({\widetilde Z}_t,t)$, so the left side of that equation is literally $\mathbb E\|\dot {\widetilde Z}_t\|^2$.  

Integrating both sides over $t\in[0,1]$ gives 
\begin{equation*}
  \int_0^1\mathbb E\|v^*(Z_t,t)\|^2\,dt \;=\; \int_0^1\mathbb E\|\dot {\widetilde Z}_t\|^2\,dt  
\end{equation*}
  By the fundamental theorem of calculus, applied pathwise (for each sample realization, not just in expectation):
$${\widetilde Z}_1-{\widetilde Z}_0 = \int_0^1 \dot {\widetilde Z}_t\,dt$$
For convenience, Define the per-sample mean velocity and deviation:
$$Y := {\widetilde Z}_1-{\widetilde Z}_0,\qquad D_t := \dot {\widetilde Z}_t-Y$$
so that $\dot {\widetilde Z}_t=Y +D_t$ for every $t$.
 Integrate $D_t$ over $t$ for a single fixed sample path:
$$\int_0^1 D_t\,dt = \int_0^1\dot {\widetilde Z}_t\,dt - \int_0^1 Y\,dt = {\widetilde Z}_1-{\widetilde Z}_0 - Y = Y - Y= 0$$
 This holds exactly, for every single sample path,  not an averaged/expectation statement, a pathwise algebraic identity.
Next, we expand the squared norm and integrate, 
$$\int_0^1\mathbb E\|\dot {\widetilde Z}_t\|^2\,dt = \int_0^1\mathbb E\|Y + D_t\|^2\,dt = \int_0^1\Big[\mathbb E\|Y\|^2+2\,\mathbb E[Y^\top D_t]+\mathbb E\|D_t\|^2\Big]dt$$

Take each term separately:
$\displaystyle\int_0^1\mathbb E\|Y \|^2\,dt = \mathbb E\|Y\|^2\cdot\int_0^1 dt = \mathbb E\|Y\|^2$.
 Applying Fubini's Theorem to the cross term, we have  
$$\int_0^1\mathbb E[Y^\top D_t]\,dt = \mathbb E\Big[Y^\top\!\!\int_0^1 D_t\,dt\Big] = \mathbb E\big[Y^\top\cdot 0\big] = 0$$

Observe that  
$\displaystyle\int_0^1\mathbb E\|D_t\|^2\,dt $ is the expected squared deviation of the instantaneous velocity from the average velocity, integrated over the path. This is 
exactly the  path curvature, denoted by $ \mathcal S(\widetilde{\mathbf Z}^{\theta}) $:
$$
\mathcal S(\widetilde{\mathbf Z}^{\theta}) =\int_0^1\mathbb E\big\|\dot{\widetilde Z}_t-(\widetilde Z_1-\widetilde Z_0)\big\|^2dt$$
Summarizing, we have 
\begin{equation*}
\int_0^1\mathbb E\|v^*(Z_t,t)\|^2\,dt  =  \int_0^1\mathbb E\|\dot {\widetilde Z}_t\|^2\,dt \;=\; \mathbb E\|{\widetilde Z}_1-{\widetilde Z}_0\|^2 \;+\; \mathcal S(\widetilde{\mathbf Z}^{\theta})
\end{equation*}





Combining the above with Eq.\ref{energy}, we conclude that
$$C= \mathbb E\|Z_1-Z_0\|^2 = \underbrace{\mathbb E\|\widetilde Z_1-\widetilde Z_0\|^2}_{\text{realized transport}} + \underbrace{\mathcal S(\widetilde{\mathbf Z}^{\theta})}_{\text{curvature}} + \underbrace{\mathcal L_{\mathrm{FM}}^{\mathrm a}(v^*)}_{\text{irreducible floor}}$$

The left side, $\mathbb E\|Z_1-Z_0\|^2$, is the original coupling's transport cost for the actual data pairing. The first term on the right, $\mathbb E\|\widetilde Z_1-\widetilde Z_0\|^2$, is the ODE-induced coupling's transport cost, likely  a different joint random pair, built by solving the field forward from the same $Z_0$, that shares the original's marginal laws but is generally a cheaper, less-crossing pairing.



 \subsection{Decomposition of Source-Target Coupling Cost: The case with $v_\theta\ne  v^*$.}

 When $v_\theta\ne v^*$, $\widetilde Z_t$ and $ Z_t$ would follow different distributions (laws) in general, and  this is precisely the training and inference distributional mismatch.
It is clear that $v_\theta(\bm z,t)$ is a fixed function given $\theta$, and the process it drives:
$$\dot{\widetilde Z}_t^{\,\theta} = v_\theta\big(\widetilde Z_t^{\,\theta},t\big),\qquad \widetilde Z_0^{\,\theta}:=Z_0$$

For convenience, define  
 ${\mathcal S(\widetilde{\mathbf Z}^{\theta} )}  :=\int_0^1\mathbb E\|\dot{\widetilde Z}_t^{\,\theta}-(\widetilde Z_1^{\,\theta}-\widetilde Z_0^{\,\theta})\|^2dt$.
Along the same line as in the case  When $v_\theta = v^*$, 
we show that 
$$\int_0^1\mathbb E\big\|v_\theta\big(\widetilde Z_t^{\,\theta},t\big)\big\|^2dt \;=\; \mathbb E\big\|\widetilde Z_1^{\,\theta}-\widetilde Z_0^{\,\theta}\big\|^2 + 
\mathcal S(\widetilde{\mathbf Z}^{\theta} )
$$

Next, define the error function $e_\theta(\bm z,t) := v_\theta(\bm z,t)-v^*(\bm z,t)$.
Expand the square:
$$\mathbb E\big\|v_\theta(\widetilde Z_t^{\,\theta},t)\big\|^2 = \mathbb E\big\|v^*(\widetilde Z_t^{\,\theta},t)\big\|^2 + 2\,\mathbb E\big[v^*(\widetilde Z_t^{\,\theta},t)^\top e_\theta(\widetilde Z_t^{\,\theta},t)\big] + \mathbb E\big\|e_\theta(\widetilde Z_t^{\,\theta},t)\big\|^2$$

Recall in the case $v_\theta =  v^*$,  $\widetilde Z_t$ solve $\dot {\widetilde {Z_t}}=v^*(\widetilde Z_t,t)$ from $Z_0$, Decompose
$$v_\theta(\widetilde Z_t^{\theta},t) - 
v^*(\widetilde Z_t,t)
=\big[v_\theta(\widetilde Z_t^{\theta}, t)-v_\theta(\widetilde {Z_t},t)\big]
+ (v_\theta(\widetilde {Z_t},t) - v^*(\widetilde Z_t,t))
$$
For convenience, define $ e_\theta= v_\theta(\widetilde {Z_t},t) - v^*(\widetilde Z_t,t)
$
It follows directly from Assumption 2 that
\[
v_\theta(\widetilde Z_t^{\theta},t) - 
v^*(\widetilde Z_t,t) \le \Lambda
\|  \widetilde Z_t^{\theta}- \widetilde Z_t\|.
\]
Appealing to the Grönwall Inequality yields 
\[
\|  \widetilde Z_t^{\theta}- \widetilde Z_t\|
\le\int_0^te^{\Lambda(t-s)}\|e_\theta(\widetilde Z_s,s)\|_{L^2}ds\le\sqrt{c_\Lambda\Delta_{FM}}.
\]
Since  $\|e_\theta(\widetilde Z_s,s)\|_{L^2}=\|e_\theta(Z_s,s)\|_{L^2}$,
we conclude that
\[
\| v_\theta(\widetilde Z_t^{\theta},t) - 
v^*(\widetilde Z_t,t) \|_{L^2} \le\Lambda\sqrt{c_\Lambda\Delta_{FM}}+\sqrt{\Delta_{FM}}=\kappa\sqrt{\Delta_{FM}}.
\]

Define 
$r_t=v_\theta(Z_t,t)-G$, and 
rewrite $W_t=[v_\theta(\widetilde Z_t,t)-v_\theta(Z_t,t)]+r_t$. We then have
\[
\|W\|_{L^2} \le\Lambda\|E\|_{L^2} + \|r\|_{L^2} \le\bar\kappa\sqrt{\mathcal L^a_{FM}},
\]
 which implies
\[
\mathcal S(\widetilde{\mathbf Z}^{\theta}) \;\le\;\bar\kappa^2\,\mathcal L^a_{FM}-
\|\widetilde Z^{\theta}_1-Z_1\|_{L_2}^2
 \;\le\;\bar\kappa^2\,\mathcal L^a_{FM}-A_0\,
 \]

 where $A_0=\mathbb E\operatorname{tr}\operatorname{Var}(Z_1\mid Z_0)$. The second step holds because $\widetilde Z^{\theta}_1$ is a function of $Z_0$.
 
Recall that  $\Delta_{\mathrm{FM}}(\theta)=\mathcal L_{\mathrm{FM}}^{\mathrm a}(\theta) -\mathcal L_{\mathrm{FM}}^{\mathrm a}(v^*)$ and  $ \mathcal L_{\mathrm{FM}}^{\mathrm a}(v^*) \le C-W_2^2$.
We conclude 
$$ \mathcal S(\widetilde{\mathbf Z}^{\theta}) 
\;\le\;\bar\kappa^2\big(  T_e +\Delta_{FM}\big)-\|\widetilde Z^{\theta}_1-Z_1\|_{L_2}^2 \,$$

\begin{corollary}
Under Assumption~\ref{assump:ode}, we have 
$$  \|     Z_1 - \widetilde Z_1^{\theta}\|_{L^2}
\le\sqrt{c_\Lambda\,\mathcal L_{\mathrm{FM}}^{\mathrm a}}, $$
where $c_\Lambda:=(e^{2\Lambda}-1)/(2\Lambda) \ge 1$.
\end{corollary}

 For convenience, define  $e_\theta:=v_\theta-v^*$, and  $G=Z_1-Z_0$.
 Let $E_t=\widetilde Z_t^\theta-Z_t$, so $E_0=0$. Since $\dot Z_t=G=v_\theta(Z_t,t)-r_t$ with $r_t=v_\theta(Z_t,t)-G$,
\begin{align*}
\dot E_t=\big[v_\theta(\widetilde Z_t^\theta,t)-v_\theta(Z_t,t)\big]+r_t\; \\
\Rightarrow\;\|E_t\|\le\int_0^te^{\Lambda(t-s)}\|r_s\|\,ds.
\end{align*}
Appealing to  Minkowski's Integral Inequality and Cauchy--Schwarz's Inequality in $s$ gives
$$  \|     Z_1 - \widetilde Z_1^{\theta}\|_{L^2}
\le 
\sqrt{c_\Lambda\,\mathcal L_{\mathrm{FM}}^{\mathrm a}}. $$

Combining the above, we conclude that smaller FM loss and $L_1$ reconstruction errors   tighten the upper bound on the coupling cost $\mathbb E\|Z_1-Z_0\|^2$ for fixed transport terms and decoder conditioning.
Since  manifold alignment (measured by the source-target coupling cost) serves as an upper  bound on the sum of path curvature and regression ambiguity of the ideal flow, better alignment between the  source and target would shape the transport map toward an easy-to-represent  function, and hence improve the performance of multi-step unrolled training.
 The FLA mechanism in VITA-WAM serves as a regularizer to drive source-target latent manifold alignment,  by ``forcing'' the decoder to reconstruct real actions from both the encoder's output and the flow's rollout endpoint:  It implicitly pressures the geometry of the two latent distributions toward each other, not just fitting better functions on top of whatever geometry already exists. 

 When the source and target are well aligned, the coupling cost between $Z_1$ and $Z_0$ is small. Euler's global error is governed by how much the velocity changes along the trajectory,  and this is intimately related to the variation that the path curvature term $\mathcal S(\widetilde{\mathbf Z})$ measures.   A smaller coupling cost tightens the upper bound on path curvature, and a good flow matching model results in small $\mathcal L_{\mathrm{FM}}^{\mathrm a}(v^*)$, making it possible that using fewer Euler steps  in solving the ODE can work well and hence reduce the inference latency.

\section{Extended Related Work}

\paragraph{Efficient World Action Models.}
WAMs have shown great promise in visuomotor control by jointly modeling world dynamics and actions~\citep{du2023unipi,feng2025vidar,pai2025mimicvideo,li2026lingbot,zhu2025uwm,bi2026motus,ye2026dreamzero,yuan2026fastwam}. State-of-the-art WAMs generally utilize large video foundation models, and iteratively generate videos and actions, incurring substantial time and space complexity and limiting their real-time deployment. The efficiency of WAMs has received extensive attention in recent studies~\citep{yuan2026fastwam,li2026efficientwam,ma2026fasterwam,zhao2026fasterwamfuture}. Fast-WAM~\citep{yuan2026fastwam} eliminates explicit future generation at inference. Faster-WAM (SparseMoT)~\citep{zhao2026fasterwamfuture} improves efficiency through selective video-action interactions and feature reuse. Efficient-WAM~\citep{li2026efficientwam} reduces the cost of future generation through compact video modeling and sparse future tokens. LaWAM~\citep{chen2026lawam} and MoWAM~\citep{wang2026mowam} leverage compact latent visual subgoals and structured motion representations, respectively. ImageWAM utilizes image-editing models to provide predictive representations for action generation~\citep{zhang2026imagewam}.  Light-WAM~\citep{li2026lightwam} employs lightweight state fusion for direct action prediction, while Faster-WAM (DoT)~\citep{ma2026fasterwam} reduces the computational cost of action generation through a shallow action expert.  Another direction accelerates iterative sampling through distillation, as demonstrated by Flash-WAM~\citep{akbari2026flashwam}. These approaches primarily improve efficiency by reducing future modeling overhead, simplifying action architectures, or accelerating sampling. A complementary direction is to reconsider the source of flow matching itself, to simplify the underlying transport required for action generation, which is discussed below.

\clearpage
\section {Ablation}
\label{sec:cond-ablation}

\subsection{Ablation of Visual Conditioning}
\label{sec:ablation-cond-ablation}

\Cref{tab:libero_visual_kv} and~\Cref{tab:libero_plus_visual_kv} compare the proposed conditioning-free VITA-WAM with VITA-WAM using visual conditioning on LIBERO and LIBERO-Plus, respectively. We report the visual-conditioning variant with $K\in\{1,2\}$ action-flow ODE integration steps.

\begin{figure*}[t]
    \centering
    \includegraphics[width=\textwidth]{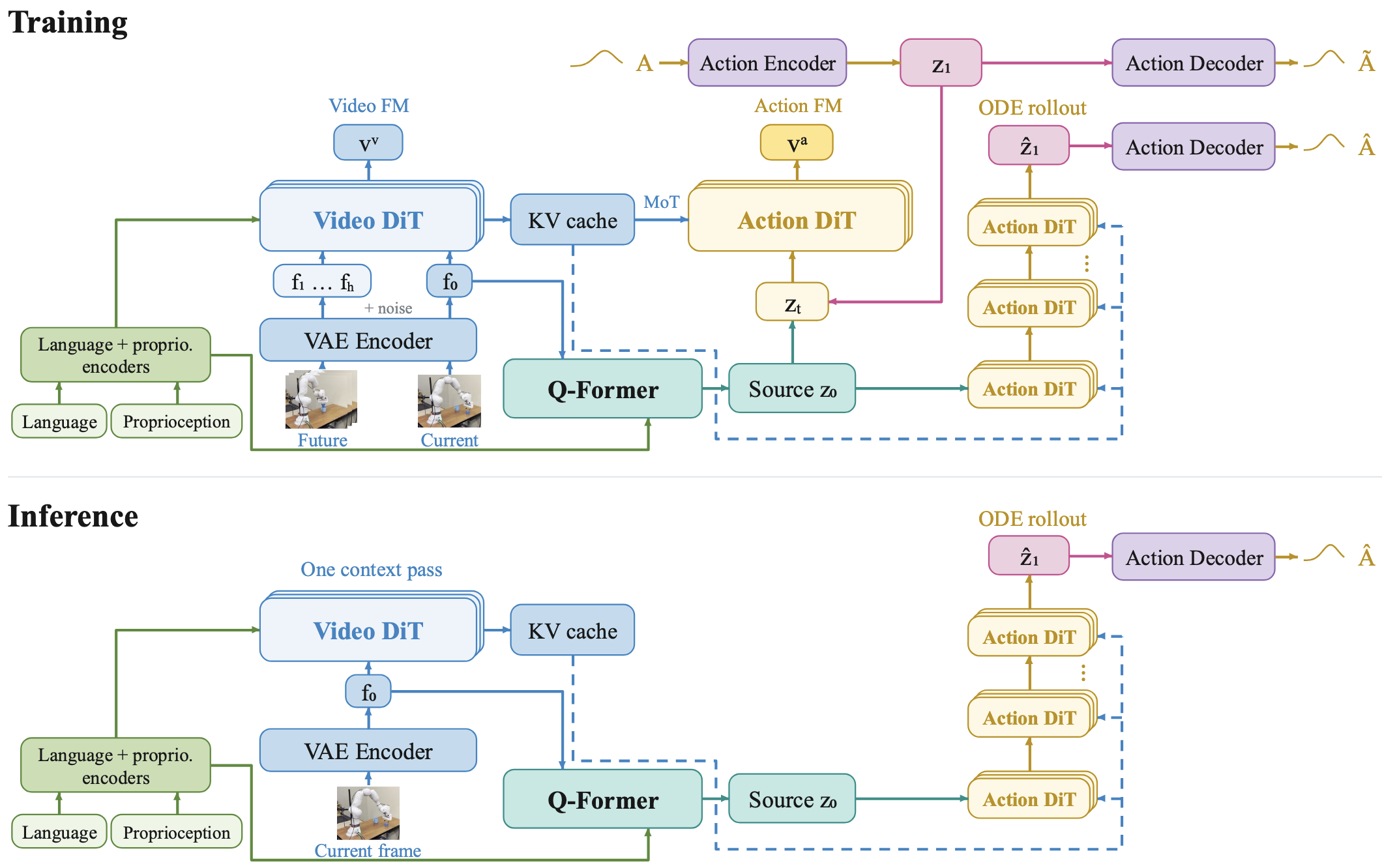}
    \caption{Training and inference architecture of VITA-WAM (visual conditioning). The Q-Former constructs the source latent from visual observations, language, and proprioception. During ODE integration, the action DiT additionally uses mixed video--action attention with cached video K/V. The action decoder converts the generated latent into an action chunk.}
    \label{fig:visual_conditioning_architecture}
\end{figure*}

\paragraph{Architecture Variants.}
Figure~\ref{fig:visual_conditioning_architecture} illustrates the visual-conditioning variant. The video backbone, text encoder, proprioceptive encoding, and multimodal source construction follow the main architecture. A Q-Former fuses visual features, language embeddings, and proprioceptive states into the source latent $\boldsymbol{z}_0$, which initializes the action flow.

The action DiT uses mixed attention over video and action tokens in a MoT architecture similar to \citep{yuan2026fastwam}. Visual keys and values are computed by the video backbone and cached for reuse at each ODE integration step. Separate language and proprioception cross-attention is removed, with these modalities incorporated through $\boldsymbol{z}_0$. The action decoder maps the generated latent $\hat{\boldsymbol{z}}_1$ to an action chunk. We denote this variant as \emph{VITA-WAM (visual conditioning)}.

\begin{figure*}[!htbp]
    \centering
    \includegraphics[width=\linewidth]{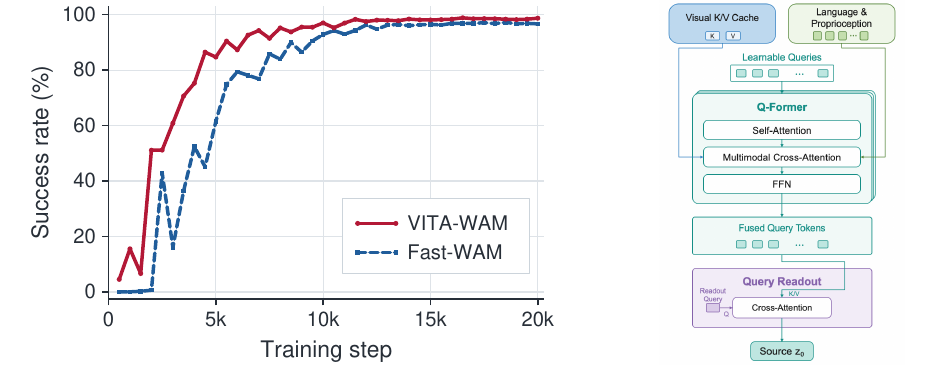}
    \par
    \begin{minipage}[t]{0.62\textwidth}
        \caption{\textbf{Training convergence on LIBERO.} Each point reports success across 40 tasks with 50 episodes per task. We evaluate both policies every 500 optimizer updates.}
        \label{fig:libero_learning_curves}
    \end{minipage}\hfill
    \begin{minipage}[t]{0.34\textwidth}
        \caption{\textbf{Q-Former source encoder.} A learned readout query pools the 16 fused query tokens into the source latent $\boldsymbol{z}_0$.}
        \label{fig:qformer_source_encoder}
    \end{minipage}
\end{figure*}

\paragraph{Training Convergence.}
Figure~\ref{fig:libero_learning_curves} compares the LIBERO training curves of VITA-WAM ($K=1$), which uses deterministic action flow, and Fast-WAM. VITA-WAM first exceeds 90\% success after 5{,}500 optimizer updates, whereas Fast-WAM first exceeds this threshold after 9{,}500 updates ($1.73\times$ as many). At the final evaluation after 20{,}000 updates, VITA-WAM achieves 98.70\% success and Fast-WAM achieves 96.65\%.

\begin{table*}[t]
    \centering
    \caption{Ablation of visual conditioning on LIBERO (SRs, \%).}
    \label{tab:libero_visual_kv}
    \small
    \setlength{\tabcolsep}{4pt}
    \begin{tabular*}{\textwidth}{@{\extracolsep{\fill}}lcccccc@{}}
        \toprule
        Method & $K$ & Spatial & Object & Goal & Long & Overall \\
        \midrule
        Fast-WAM & 10 & 98.20 & 99.60 & 97.60 & 93.40 & 97.20 \\
        \textbf{VITA-WAM} & 1 & \textbf{100.00} & 99.60 & 98.20 & \textbf{97.60} & \textbf{98.85} \\
        VITA-WAM (visual conditioning) & 1 & 98.80 & \textbf{100.00} & 98.20 & 95.80 & 98.20 \\
        VITA-WAM (visual conditioning) & 2 & 97.80 & 99.60 & 98.20 & 97.20 & 98.20 \\
        VITA-WAM (visual conditioning) & 4 & 97.40 & 99.60 & \textbf{98.80} & 96.40 & 98.05 \\
        \bottomrule
    \end{tabular*}
\end{table*}

\begin{table*}[t]
    \centering
    \caption{Ablation of visual conditioning on LIBERO-Plus (SRs, \%).}
    \label{tab:libero_plus_visual_kv}
    \footnotesize
    \setlength{\tabcolsep}{2.5pt}
    \begin{tabular*}{\textwidth}{@{\extracolsep{\fill}}lccccccccc@{}}
        \toprule
        Method & $K$ & Camera & Robot & Lang. & Light & BG & Noise & Layout & Overall \\
        \midrule
        Fast-WAM & 10 & 10.01 & 46.90 & 67.66 & 82.57 & \textbf{56.78} & 28.04 & 66.75 & 49.33 \\
        \textbf{VITA-WAM} & 1 & \textbf{57.72} & \textbf{67.48} & 85.69 & 88.53 & 54.18 & \textbf{64.21} & \textbf{78.23} & \textbf{70.80} \\
        VITA-WAM (visual conditioning)$^{\dagger}$ & 1 & 37.65 & 59.16 & 90.70 & 89.05 & 48.98 & 48.66 & 74.23 & 63.49 \\
        VITA-WAM (visual conditioning) & 2 & 31.71 & 61.68 & \textbf{91.41} & \textbf{89.49} & 46.10 & 37.16 & 74.30 & 60.96 \\
        VITA-WAM (visual conditioning) & 4 & 29.71 & 56.19 & 91.22 & 87.48 & 48.42 & 39.79 & 73.84 & 60.13 \\
        \bottomrule
    \end{tabular*}
\end{table*}

\section{Real-World Tasks}
\label{app:real_world_data}

\begin{table*}[htbp]
    \centering
    \caption{Real-world demonstration tasks, instructions, and task outcomes.}
    \label{tab:real_demo_tasks}
    \small
    \begin{tabular}{@{}l p{0.34\textwidth} p{0.40\textwidth}@{}}
        \toprule
        Task & Instruction & Outcome \\
        \midrule
        Pick Place & Pick up the blue cube and place it into the light blue mug. & The cube rests inside the upright mug. \\
        \addlinespace
        Stack Cups & Stack teal cup to orange cup.\par Stack orange cup to teal cup. & The specified upper cup sits upright inside the receiving cup after release. \\
        \addlinespace
        Hang Mug & Hang the dark blue mug by its handle on the wooden stand.\par Hang the light blue mug by its handle on the wooden stand. & The mug hangs by its handle from the peg after release. \\
        \addlinespace
        Store Drawer & Put the blue four-lobed block in the drawer and close it.\par Put the red polygonal block in the drawer and close it. & The block is inside the box and the sliding lid covers the opening. \\
        \bottomrule
    \end{tabular}
\end{table*}

\begin{figure*}[!htbp]
    \centering
    \includegraphics[width=\textwidth]{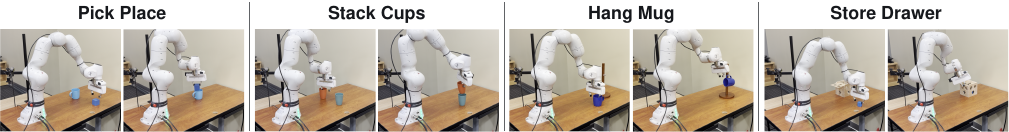}
    \caption{Real-world tasks on the Franka Research 3. Each pair shows two stages of the same task in temporal order.}
    \label{fig:task_viz}
\end{figure*}

\paragraph{Demonstration collection.}
An operator teleoperates a Franka Research 3 arm equipped with a Franka Hand using a Meta Quest controller. Three ZED cameras record RGB video at a nominal 30\,Hz, one from the wrist and two from external viewpoints. Each demonstration pairs a language instruction and the three video streams with a sequence of robot states and actions.
The actions have eight dimensions, comprising seven absolute joint-position targets in radians and one binary gripper command.
We retain 180 successful demonstrations per task to train VITA-WAM ($K=1$), for a total of 720 demonstrations across the four tasks in Figure~\ref{fig:task_viz}. Table~\ref{tab:real_demo_tasks} describes the instructions and the outcomes observed during demonstration review.

\paragraph{Training example construction.}
At each sampled time point, we crop the two external camera images, resize each to $112\times224$ pixels, and stack them vertically. We resize the full wrist image to $224\times224$ pixels and place it to their right, yielding a $224\times448$ RGB image (height $\times$ width). We draw training windows from 33 consecutive frames within a single uninterrupted segment of teleoperation. The instruction, initial image, and initial robot state condition a chunk of $H=32$ consecutive eight-dimensional action targets. We normalize the action targets using per-channel minima and maxima from the training demonstrations. Following \href{https://arxiv.org/abs/2603.16666}{Fast-WAM}, we supervise the video prediction with nine images sampled at frame offsets $0,4,\ldots,32$.

\section{Inference Latency}
\label{app:inference_latency}

\paragraph{Evaluating Efficiency.}
We benchmark action generation at batch size one on a single NVIDIA RTX 4090 for all policies. Each timing starts with three RGB images, the robot state, and the language instruction. We measure wall-clock time through image preprocessing, tokenization, policy inference, and transfer of the predicted action chunk to the CPU. For policies that use UMT5, each call tokenizes the instruction and runs the text encoder once.
For each configuration, we run 200 warm-up calls and report the median of 2000 subsequent calls in ~\Cref{tab:inference_efficiency}.

\section{Implementation Details}
\label{app:implementation}

The following settings describe the conditioning-free VITA-WAM policies trained on LIBERO and evaluated on LIBERO-Plus without additional training.

\subsection{Model Architecture}
\label{app:implementation_architecture}

\paragraph{Backbone.}
We use the pretrained Wan2.2-TI2V-5B video backbone, together with its image VAE and text encoder, following \href{https://arxiv.org/abs/2603.16666}{Fast-WAM}. The action transformer has 30 layers with hidden width 1024 and FFN width 4096. Its attention projections retain 24 heads of dimension 128. We initialize the action backbone from Wan2.2-derived, width-interpolated ActionDiT weights, while its input and output projections, the source encoder, and the action autoencoder are newly initialized.

\paragraph{Source encoder.}
Figure~\ref{fig:qformer_source_encoder} shows how the Q-Former combines visual, language, and proprioceptive inputs into the source latent $\boldsymbol{z}_0$.
The Q-Former constructs the source from the final video layer's current-frame K/V, instruction embeddings, and proprioception. Separate linear adapters project visual keys and values from width 3072 to 512, while language and proprioception are projected to the same width. The Q-Former contains four blocks with 16 learned queries, eight attention heads, and FFN width 2048. Each block applies query self-attention, language/proprioception cross-attention, visual cross-attention, and an FFN sequentially. A learned readout query aggregates the resulting queries into one 512-dimensional source token. The current frame remains clean during training, and a first-frame causal mask prevents its features from accessing future observations. At inference, one current-frame video pass at zero video noise supplies the source features.

\paragraph{Action transformer and action autoencoder.}
ActionDiT maps the 512-dimensional flow state to hidden width 1024 and predicts a 512-dimensional velocity. Its blocks operate only on the evolving action token, with no language/proprioception cross-attention or video K/V conditioning.  We initialize the blocks from the Wan2.2-derived, width-interpolated ActionDiT checkpoint and train the input projection and velocity head from scratch.
The action autoencoder maps between a $32\times7$ action chunk and a single 512-dimensional latent token. Its encoder uses four Conv1d--ReLU stages with 512 channels, kernel size five, and stride two, followed by a linear projection. The decoder first maps the latent token to two time positions, each with 256 features. Four upsampling stages expand the sequence to 4, 8, 16, and finally 32 positions. A final projection predicts seven action coordinates at each position. We use the same decoder to reconstruct demonstrations from $\boldsymbol{z}_1$ and to predict actions from the flow endpoint $\hat{\boldsymbol{z}}_1$ during FLA training and inference.

\subsection{Training Details}
\label{app:implementation_training}

\paragraph{Joint training.}
We jointly optimize the video backbone, ActionDiT, source encoder, proprioception projection, and action autoencoder. The image VAE remains frozen, and instruction embeddings are precomputed. Action losses propagate through the source encoder into the video backbone. The video FM, action FM, AE, and FLA objectives have weights $1$, $1$, $0.5$, and $0.25$, respectively.

\paragraph{FM and FLA Training.}
Video and action flow times are sampled independently. Action FM samples time uniformly on $[0,1]$ and uses a timestep-weighted velocity loss; the video scheduler retains shift 5. For FLA, we construct the source once and integrate it using $K$ explicit Euler steps on a uniform grid from source time $t=0$ to target time $t=1$. The endpoint is decoded and supervised with an L1 action loss, with gradients propagated through the integration steps and decoder. AE and FLA losses are averaged over valid action timesteps, excluding padding. FLA is applied on every training update from the start of training. Each configuration uses the same $K$ during training and inference, with all other settings shared.

\paragraph{Optimization.}
We follow Fast-WAM's data preparation and use training-set statistics to normalize actions and proprioception. The external and wrist views are concatenated horizontally into a $224\times448$ input. Each training sample contains a 32-step action chunk and nine video frames sampled at offsets $0,4,\ldots,32$. We train jointly on the four LIBERO suites for 20,000 optimizer updates on two NVIDIA H200 GPUs, using BF16 and DeepSpeed ZeRO-1. All trainable modules share one AdamW optimizer and learning-rate schedule. Table~\ref{tab:implementation_training} summarizes the training configuration.

\begin{table}[htbp]
    \centering
    \caption{\textbf{LIBERO training configuration.} The global batch size is $8$ examples per GPU $\times\,2$ GPUs $\times\,8$ gradient-accumulation steps. Schedule durations refer to optimizer updates.}
    \label{tab:implementation_training}
    \small
    \renewcommand{\arraystretch}{1.08}
    \begin{tabular}{@{}p{0.43\textwidth}p{0.52\textwidth}@{}}
        \toprule
        Setting & Value \\
        \midrule
        Training hardware & 2 NVIDIA H200 GPUs \\
        Distributed strategy / precision & DeepSpeed ZeRO-1 / BF16 \\
        Optimizer & AdamW; $\beta=(0.9,0.95)$ \\
        Peak / minimum learning rate & $10^{-4}$ / $10^{-6}$ \\
        Weight decay & $0.01$ \\
        Learning-rate schedule & 1,000-update linear warmup, then cosine decay \\
        Training budget & 20,000 optimizer updates \\
        Global batch size & 128 \\
        Gradient-norm limit & 1.0 \\
        Video / action scheduler shift & 5 / 1 \\
        Video FM / action FM / AE / FLA weights & $1$ / $1$ / $0.5$ / $0.25$ \\
        Training seed & 42 \\
        \bottomrule
    \end{tabular}
\end{table}

\end{document}